\documentclass[10pt,journal,compsoc]{IEEEtran}
\usepackage[nocompress]{cite}

\usepackage{color}
\usepackage{soul}
\sethlcolor{white}
\usepackage{rotating}
\usepackage{arydshln}
\usepackage{stackengine} 
\stackMath
\usepackage{enumitem}
\usepackage{amsmath}
\usepackage{amssymb}
\usepackage{newunicodechar}
\usepackage{bm} 

\begin{document}
\title{A General Survey on Attention Mechanisms in Deep Learning}

\author{Gianni~Brauwers\IEEEmembership{}
        and~Flavius~Frasincar\IEEEmembership{}
\thanks{G. Brauwers and F. Frasincar are with the Erasmus School of Economics, Erasmus University Rotterdam, 3000 DR, Rotterdam, the Netherlands (e-mail: \{frasincar, brauwers\}@ese.eur.nl).}
\thanks{Manuscript received July 6, 2020; revised June 21, 2021;
Corresponding author: F. Frasincar}}


\IEEEtitleabstractindextext{%
\begin{abstract}
Attention is an important mechanism that can be employed for a variety of deep learning models across many different domains and tasks. This survey provides an overview of the most important attention mechanisms proposed in the literature. The various attention mechanisms are explained by means of a framework consisting of a general attention model, uniform notation, and a comprehensive taxonomy of attention mechanisms. Furthermore, the various measures for evaluating attention models are reviewed, and methods to characterize the structure of attention models based on the proposed framework are discussed. Last, future work in the field of attention models is considered.
\end{abstract}
}

\maketitle

\IEEEdisplaynontitleabstractindextext

\IEEEpeerreviewmaketitle

\IEEEraisesectionheading{\section{Introduction}\label{sec:introduction}}

\IEEEPARstart{T}{he} idea of mimicking human attention first arose in the field of computer vision \cite{larochelle2010learning, mnih2014recurrent} in an attempt to reduce the computational complexity of image processing while improving performance by introducing a model that would only focus on specific regions of images instead of the entire picture. Although, the true starting point of the attention mechanisms we know today is often attributed to originate in the field of natural language processing \cite{bahdanau2014neural}. Bahdanau et al. \cite{bahdanau2014neural} implement attention in a machine translation model to address certain issues with the structure of recurrent neural networks. After Bahdanau et al. \cite{bahdanau2014neural} emphasized the advantages of attention, the attention techniques were refined \cite{luong2015effective} and quickly became popular for a variety of tasks, such as text classification \cite{yang-etal-2016-hierarchical, wang-etal-2016-attention}, image captioning \cite{anderson2018bottom, xu2015show}, sentiment analysis \cite{wang-etal-2016-attention, ma2018targeted}, and speech recognition \cite{chorowski2015attention, bahdanau2016end, kim2017joint}.

Attention has become a popular technique in deep learning for several reasons. Firstly, models that incorporate attention mechanisms attain state-of-the-art results for all of the previously mentioned tasks, and many others. Furthermore, most attention mechanisms can be trained jointly with a base model, such as a recurrent neural network or a convolutional neural network using regular backpropagation \cite{bahdanau2014neural}. Additionally, attention introduces a certain type of interpretation into neural network models \cite{xu2015show} that are generally known to be highly complicated to interpret. Moreover, the popularity of attention mechanisms was additionally boosted after the introduction of the Transformer model \cite{vaswani2017attention} that further proved how effective attention can be. Attention was originally introduced as an extension to recurrent neural networks \cite{cho2014properties}. However, the Transformer model proposed in \cite{vaswani2017attention} poses a major development in attention research as it demonstrates that the attention mechanism is sufficient to build a state-of-the-art model. This means that disadvantages, such as the fact that recurrent neural networks are particularly difficult to parallelize, can be circumvented. As was the case for the introduction of the original attention mechanism \cite{bahdanau2014neural}, the Transformer model was created for machine translation, but was quickly adopted to be used for other tasks, such as image processing \cite{parmar2018image}, video processing \cite{zhou2018end}, and recommender systems \cite{sun2019bert4rec}.

The purpose of this survey is to explain the general form of attention, and provide a comprehensive overview of attention techniques in deep learning. Other surveys have already been published on the subject of attention models. For example, in \cite{wang2016survey}, a survey is presented on attention in computer vision, \cite{lee2018attention} provides an overview of attention in graph models, and \cite{chaudhari2019attentive, hu2018introductory, galassi2019attention} are all surveys on attention in natural language processing. This paper partly builds on the information presented in the previously mentioned surveys. Yet, we provide our own significant contributions. The main difference between this survey and the previously mentioned ones is that the other surveys generally focus on attention models within a certain domain. This survey, however, provides a cross-domain overview of attention techniques. We discuss the attention techniques in a general way, allowing them to be understood and applied in a variety of domains. Furthermore, we found the taxonomies presented in previous surveys to be lacking the depth and structure needed to properly distinguish the various attention mechanisms. Additionally, certain significant attention techniques have not yet been properly discussed in previous surveys, while other presented attention mechanisms seem to be lacking either technical details or intuitive explanations. Therefore, in this paper, we present important attention techniques by means of a single framework using a uniform notation, a combination of both technical and intuitive explanations for each presented attention technique, and a comprehensive taxonomy of attention mechanisms.

The structure of this paper is as follows. Section \ref{sec:GeneralAttention} introduces a general attention model that provides the reader with a basic understanding of the properties of attention and how it can be applied. One of the main contributions of this paper is the taxonomy of attention techniques presented in Section \ref{sec:Taxonomy}. In this section, attention mechanisms are explained and categorized according to the presented taxonomy. Section \ref{sec:Evaluation} provides an overview of performance measures and methods for evaluating attention models. Furthermore, the taxonomy is used to evaluate the structure of various attention models. Lastly, in Section \ref{sec:Conclusion}, we give our conclusions and suggestions for further research.

\begin{figure}
    \centering
    \includegraphics[scale=0.55]{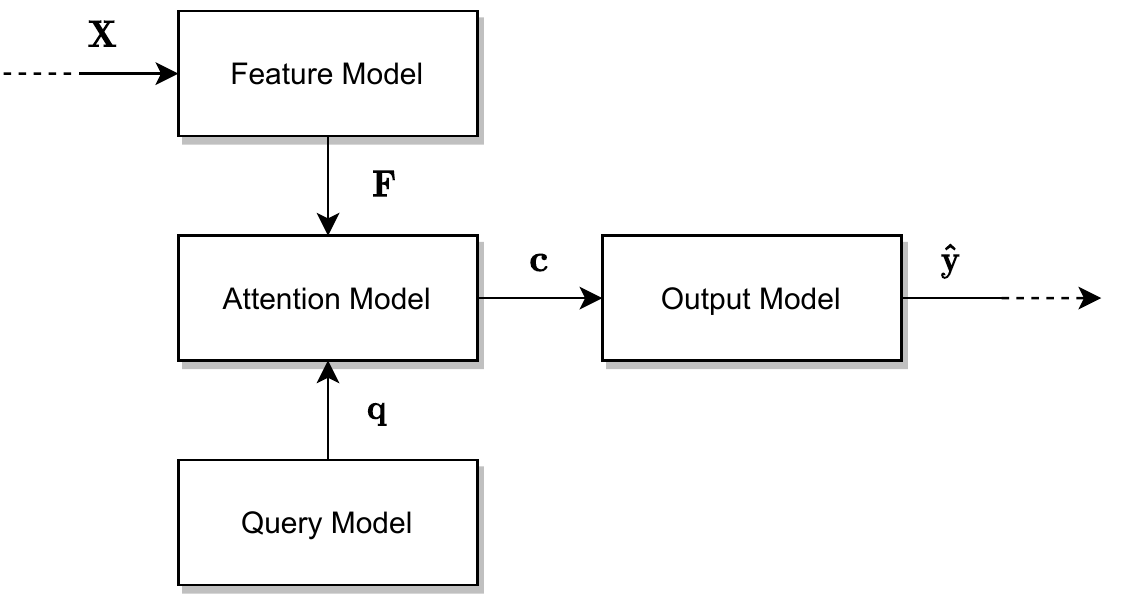}
    \caption{An illustration of the general structure of the task model.}
    \label{fig:TaskModel}
\end{figure}

\section{General Attention Model}\label{sec:GeneralAttention}

This section presents a general form of attention with corresponding notation. The notation introduced here is based on the notation that was introduced in \cite{daniluk2017frustratingly} and popularized in \cite{vaswani2017attention}. The framework presented in this section is used throughout the rest of this paper.

To implement a general attention model, it is necessary to first describe the general characteristics of a model that can employ attention. First of all, we will refer to the complete model as the \textbf{task model}, of which the structure is presented in Fig. \ref{fig:TaskModel}. This model simply takes an input, carries out the specified task, and produces the desired output. For example, the task model can be a language model that takes as input a piece of text, and produces as output a summary of the contents, a classification of the sentiment, or the text translated word for word to another language. Alternatively, the task model can take an image, and produce a caption or segmentation for that image. The task model consists of four submodels: the feature model, the query model, the attention model, and the output model. In Subsection \ref{sec:AttentionInput}, the feature model and query model are discussed, which are used to prepare the input for the attention calculation. In Subsection \ref{sec:AttentionOutput}, the attention model and output model are discussed, which are concerned with producing the output.

\subsection{Attention Input}\label{sec:AttentionInput}

Suppose the task model takes as input the matrix $\bm X \in \mathbb{R}^{d_x \times n_x}$, where $d_x$ represents the size of the input vectors and $n_x$ represents the amount of input vectors. The columns in this matrix can represent the words in a sentence, the pixels in an image, the characteristics of an acoustic sequence, or any other collection of inputs. The \textbf{feature model} is then employed to extract the $n_f$ feature vectors $\bm f_1, \dots, \bm f_{n_f} \in \mathbb{R}^{d_f}$ from $\bm X$, where $d_f$ represents the size of the feature vectors. The feature model can be a recurrent neural network (RNN), a convolutional neural network (CNN), a simple embedding layer, a linear transformation of the original data, or no transformation at all. Essentially, the feature model consists of all the steps that transform the original input $\bm X$ into the feature vectors $\bm f_1, \dots, \bm f_{n_f}$ that the attention model will attend to.

To determine which vectors to attend to, the attention model requires the query $\bm q \in \mathbb{R}^{d_q}$, where $d_q$ indicates the size of the query vector. This query is extracted by the \textbf{query model}, and is generally designed based on the type of output that is desired of the model. A query tells the attention model which feature vectors to attend to. It can be interpreted literally as a query, or a question. For example, for the task of image captioning, suppose that one uses a decoder RNN model to produce the output caption based on feature vectors obtained from the image by a CNN. At each prediction step, the hidden state of the RNN model can be used as a query to attend to the CNN feature vectors. In each step, the query is a question in the sense that it asks for the necessary information from the feature vectors based on the current prediction context.

\begin{figure}
    \centering
    \includegraphics[scale=0.6]{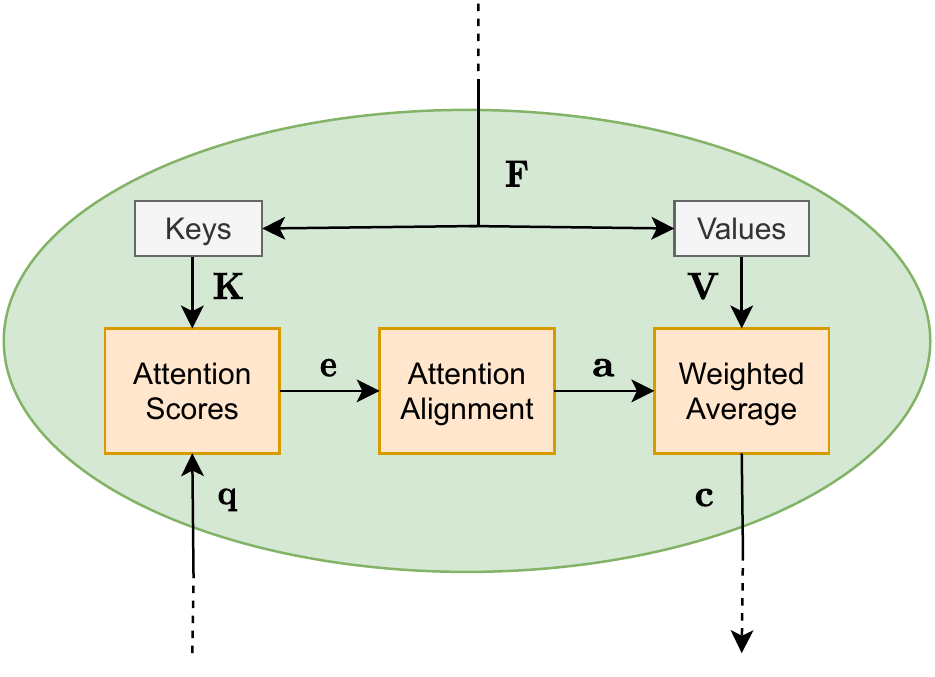}
    \caption{The inner mechanisms of the general attention module.}
    \label{fig:GeneralAttentionModule}
\end{figure}

\subsection{Attention Output}\label{sec:AttentionOutput}

The feature vectors and query are used as input for the \textbf{attention model}. This model consists of a single, or a collection of \textbf{general attention modules}. An overview of a general attention module is presented in Fig. \ref{fig:GeneralAttentionModule}. The input of the general attention module is the query $\bm q \in \mathbb{R}^{d_{q}}$, and the matrix of feature vectors $\bm F = [\bm f_1, \dots , \bm f_{n_f}] \in \mathbb{R}^{d_{f} \times n_f}$. Two separate matrices are extracted from the matrix $\bm F$: the keys matrix $\bm K = [\bm k_1, \dots , \bm k_{n_f}] \in \mathbb{R}^{d_k \times n_f}$, and the values matrix $\bm V = [\bm v_1, \dots, \bm v_{n_f}] \in \mathbb{R}^{d_v \times n_f}$, where $d_k$ and $d_v$ indicate, respectively, the dimensions of the key vectors (columns of $\bm K$) and value vectors (columns of $\bm V$). The general way of obtaining these matrices is through a linear transformation of $\bm F$ using the weight matrices $\bm W_K \in \mathbb{R}^{d_k \times d_f}$ and $\bm W_V \in \mathbb{R}^{d_v \times d_f}$, for $\bm K$ and $\bm V$, respectively. The calculations of $\bm  K$ and $\bm V$ are presented in (\ref{equation:Keys&Values}). Both weight matrices can be learned during training or predefined by the researcher. For example, one can choose to define both $\bm W_K$ and $\bm W_V$ as equal to the identity matrix to retain the original feature vectors. Other ways of defining the keys and the values are also possible, such as using completely separate inputs for the keys and values. The only constraint to be obeyed is that the number of columns in $\bm K$ and $\bm V$ remains the same.
\begin{equation}\label{equation:Keys&Values}
    \def\sss{\scriptscriptstyle}
    \setstackgap{L}{8pt}
    \def\stacktype{L}
    \stackunder{\bm K}{\sss d_k \times n_f} = \stackunder{\bm W_K}{\sss d_k \times d_f} \times \stackunder{\bm F}{\sss d_f \times n_f},
    \hspace{20pt}
    \stackunder{\bm V}{\sss d_v \times n_f} = \stackunder{\bm W_V}{\sss d_v \times d_f} \times \stackunder{\bm F}{\sss d_f \times n_f}.
\end{equation}

The goal of the attention module is to produce a weighted average of the value vectors in $\bm V$. The weights used to produce this output are obtained via an attention scoring and alignment step. The query $\bm q$ and the keys matrix $\bm K$ are used to calculate the vector of attention scores $\bm e = [e_{1}, \dots, e_{n_f}]\in \mathbb{R}^{n_f}$. This is done via the score function $\text{score}()$, as illustrated in (\ref{equation:GeneralAttentionScore}).
\begin{equation}\label{equation:GeneralAttentionScore}
    \def\sss{\scriptscriptstyle}
    \setstackgap{L}{8pt}
    \def\stacktype{L}
   \stackunder{e_{l}}{\sss 1 \times 1} = \text{score}(\stackunder{\bm q}{\sss d_q \times 1}, \stackunder{\bm k_l}{\sss d_k \times 1}). 
\end{equation}
As discussed before, the query symbolizes a request for information. The attention score $e_l$ represents how important the information contained in the key vector $\bm k_l$ is according to the query. If the dimensions of the query and key vectors are the same, an example of a score function would be to take the dot-product of the vectors. The different types of score functions are further discussed in Section \ref{sec:ScoreFunction}.

Next, the attention scores are processed further through an alignment layer. The attention scores can generally have a wide range outside of $[0, 1]$. However, since the goal is to produce a weighted average, the scores are redistributed via an alignment function $\text{align}()$ as defined in (\ref{equation:GeneralAttentionWeight}).
\begin{equation}\label{equation:GeneralAttentionWeight}
    \def\sss{\scriptscriptstyle}
    \setstackgap{L}{8pt}
    \def\stacktype{L}
    \stackunder{a_{l}}{\sss 1 \times 1} = \text{align}(\stackunder{e_{l}}{\sss 1 \times 1}; \stackunder{\bm e}{\sss n_f \times 1}),
\end{equation}
where $a_l \in \mathbb{R}^{1}$ is the attention weight corresponding to the $l$th value vector.  One example of an alignment function would be to use a softmax function, but the various other alignment types are discussed in Section \ref{sec:Alignment}. The attention weights provide a rather intuitive interpretation for the attention module. Each weight is a direct indication of how important each feature vector is relative to the others for this particular problem. This can provide us with a more in-depth understanding of the model behaviour, and the relations between inputs and outputs. The vector of attention weights $\bm a = [a_{1}, \dots, a_{n_f}]\in \mathbb{R}^{n_f}$ is used to produce the context vector $\bm c \in \mathbb{R}^{d_{v}}$ by calculating a weighted average of the columns of the values matrix $\bm V$, as shown in (\ref{equation:GeneralContextVector}). 
\begin{equation}\label{equation:GeneralContextVector}
    \def\sss{\scriptscriptstyle}
    \setstackgap{L}{8pt}
    \def\stacktype{L}
    \stackunder{\bm c}{\sss d_v \times 1} = \sum^{n_f}_{l=1} \stackunder{a_{l}}{\sss 1 \times 1} \times \stackunder{\bm v_l}{\sss d_v \times 1}.
\end{equation}

As illustrated in Fig. \ref{fig:TaskModel}, the context vector is then used in the \textbf{output model} to create the output $\hat{\bm y}$. This output model translates the context vector into an output prediction. For example, it could be a simple softmax layer that takes as input the context vector $\bm c$, as shown in (\ref{equation:OutputModelSoftmax}).
\begin{equation}\label{equation:OutputModelSoftmax}
    \def\sss{\scriptscriptstyle}
    \setstackgap{L}{8pt}
    \def\stacktype{L}
    \stackunder{\hat{\bm y}}{\sss d_y \times 1} = \text{softmax}( \stackunder{\bm W_c}{\sss d_y \times d_v} \times \stackunder{\bm c}{\sss d_v \times 1} +  \stackunder{\bm b_c}{\sss d_y \times 1}),
\end{equation}
where $d_y$ is the number of output choices or classes, and $\bm W_c \in \mathbb{R}^{d_y \times d_v}$ and $\bm b_c \in \mathbb{R}^{d_y}$ are trainable weights.

\subsection{Attention Applications}

Attention is a rather general mechanism that can be used in a wide variety of problem domains. Consider the task of machine translation using an RNN model. Also, consider the problem of image classification using a basic CNN model. While an RNN produces a sequence of hidden state vectors, a CNN creates feature maps, where each region in the image is represented by a feature vector. The RNN hidden states are organized sequentially, while the CNN feature maps are organized spatially. Yet, attention can still be applied in both situations, since the attention mechanism does not inherently depend on the organization of the feature vectors. This characteristic makes attention easy to implement in a wide variety of models in different domains.

Another domain where attention can be applied is audio processing \cite{xu2017attention, yu2018multi}. Acoustic sequences can be represented by a sequence of feature vectors that relate to certain time periods of the audio sample. These vectors could simply be the raw input audio, or they can be extracted via, for example, an RNN or CNN. Video processing is another domain where attention can be applied intuitively \cite{sharma2015action, gao2017video}. Video data consists of sequences of images, so attention can be applied to the individual images, as well as the entire sequence. Recommender systems often incorporate a user's interaction history to produce recommendations. Feature vectors can be extracted based on, for example, the id's or other characteristics of the products the user interacted with, and attention can be applied to them \cite{ying2018sequential}. Attention can generally also be applied to many problems that use a time series as input, be it medical \cite{song2018attend}, financial \cite{8476227}, or anything else, as long as feature vectors can be extracted.

The fact that attention does not rely on the organization of the feature vectors allows it to be applied to various problems that each use data with different structures, as illustrated by the previous domain examples. Yet, this can be taken even further by applying attention to data where there is irregular structure. For example, protein structures, city traffic flows, and communication networks cannot always be represented using neatly structured organizations, such as sequences, like time series, or grids, like images. In such cases, the different aspects of the data are often represented as nodes in a graph. These nodes can be represented by feature vectors, meaning that attention can be applied in domains that use graph-structured data as well \cite{velivckovic2017graph, lee2018attention}. 

In general, attention can be applied to any problem for which a set of feature vectors can be defined or extracted. As such, the general attention model presented in Fig. \ref{fig:GeneralAttentionModule} is applicable to a wide range of domains. The problem, however, is that there is a large variety of different applications and extensions of the general attention module. As such, in Section \ref{sec:Taxonomy}, a comprehensive overview is provided of a collection of different attention mechanisms.

\begin{figure*}
    \centering
    \includegraphics[width=0.9\linewidth]{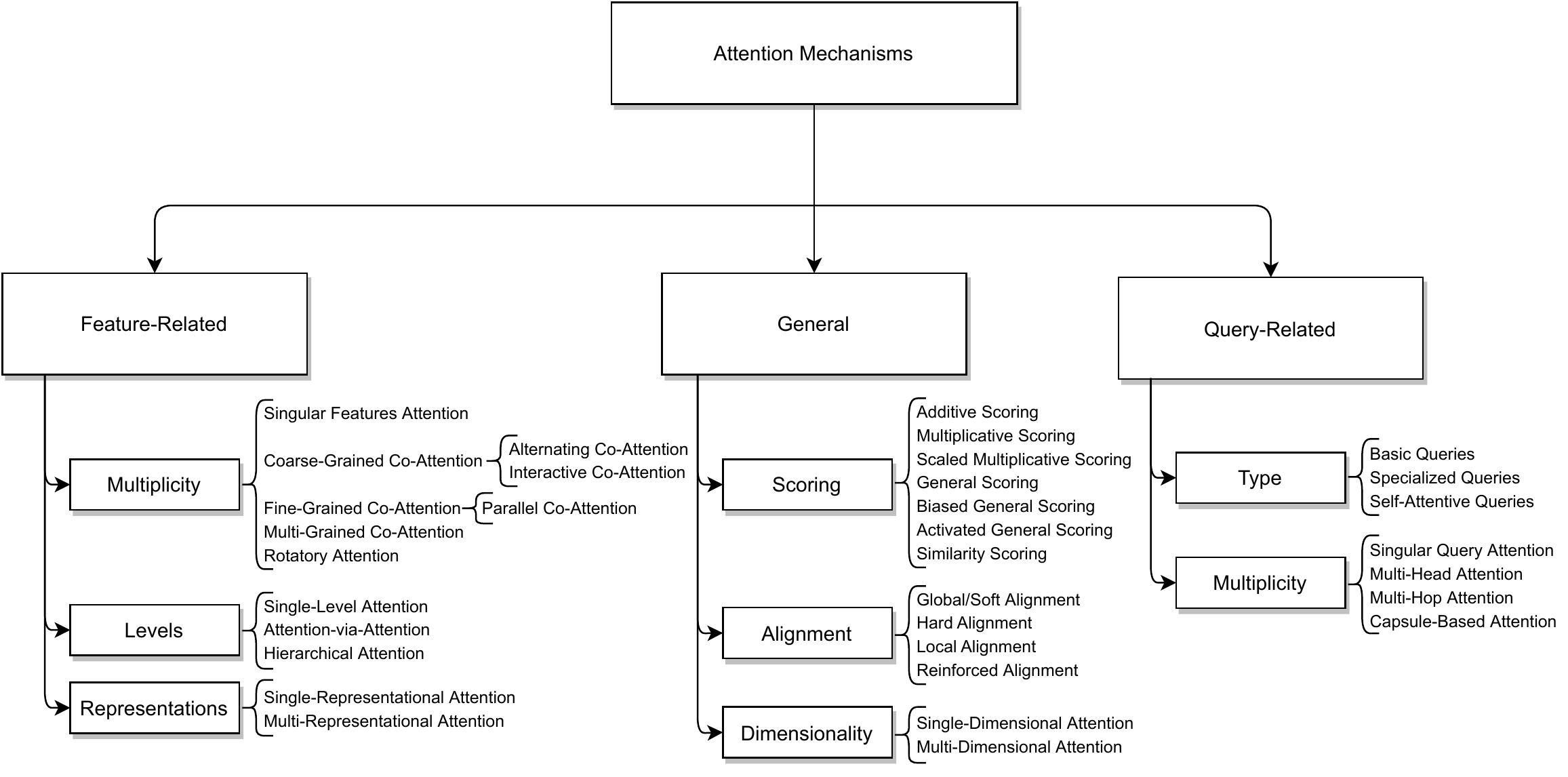}
    \caption{A taxonomy of attention mechanisms.}
    \label{fig:Taxonomy}
\end{figure*}

\section{Attention Taxonomy}\label{sec:Taxonomy}
There are many different types of attention mechanisms and extensions, and a model can use different combinations of these attention techniques. As such, we propose a taxonomy that can be used to classify different types of attention mechanisms. Fig. \ref{fig:Taxonomy} provides a visual overview of the different categories and subcategories that the attention mechanisms can be organized in. The three major categories are based on whether an attention technique is designed to handle specific types of feature vectors (feature-related), specific types of model queries (query-related), or whether it is simply a general mechanism that is related to neither the feature model, nor the query model (general). Further explanations of these categories and their subcategories are provided in the following subsections. Each mechanism discussed in this section is either a modification to the existing inner mechanisms of the general attention module presented in Section \ref{sec:GeneralAttention}, or an extension of it.

The presented taxonomy can also be used to analyze the architecture of attention models. Namely, the major categories and their subcategories can be interpreted as orthogonal dimensions of an attention model. An attention model can consist of a combination of techniques taken from any or all categories. Some characteristics, such as the scoring and alignment functions, are generally required for any attention model. Other mechanisms, such as multi-head attention or co-attention are not necessary in every situation. Lastly, in Table \ref{Table:Notation}, an overview of used notation with corresponding descriptions is provided.

\begin{table}
\caption{Notation.}
\begin{center}
\resizebox{0.99\linewidth}{!}{%
\begin{tabular}{lp{200pt}}
\hline
Symbol & Description \\ \hline
$\bm F$ & Matrix of size $d_f \times n_f$ containing the feature vectors $\bm f_1, \dots, \bm f_{n_f} \in \mathbb{R}^{d_{f}}$ as columns. These feature vectors are extracted by the feature model. \\

$\bm K$ & Matrix of size $d_k \times n_f$ containing the key vectors $\bm k_1, \dots, \bm k_{n_f} \in \mathbb{R}^{d_{k}}$ as columns. These vectors are used to calculate the attention scores. \\

$\bm V$ & Matrix of size $d_v \times n_f$ containing the value vectors $\bm v_1, \dots, \bm v_{n_f} \in \mathbb{R}^{d_{v}}$ as columns. These vectors are used to calculate the context vector. \\
    
$\bm W_K$ & Weights matrix of size $d_k \times d_f$ used to create the $\bm K$ matrix from the $\bm F$ matrix. \\
    
$\bm W_V$ & Weights matrix of size $d_v \times d_f$ used to create the $\bm V$ matrix from the $\bm F$ matrix. \\
    
$\bm q$ & Query vector of size $d_q$. This vector essentially represents a question, and is used to calculate the attention scores. \\
    
$\bm c$ & Context vector of size $d_v$. This vector is the output of the attention model. \\
    
$\bm e$ & Score vector of size $d_{n_f}$ containing the attention scores $e_{1}, \dots, e_{n_f} \in \mathbb{R}^{1}$. These are used to calculate the attention weights. \\
    
$\bm a$ & Attention weights vector of size $d_{n_f}$ containing the attention weights $a_{1}, \dots, a_{n_f} \in \mathbb{R}^{1}$. These are the weights used in the calculation of the context vector.
\end{tabular}
}
\end{center}
\label{Table:Notation}
\end{table}

\subsection{Feature-Related Attention Mechanisms}\label{sec:ModelInput}
Based on a particular set of input data, a feature model extracts feature vectors so that the attention model can attend to these various vectors. These features may have specific structures that require special attention mechanisms to handle them. These mechanisms can be categorized to deal with one of the following feature characteristics: the multiplicity of features, the levels of features, or the representations of features.

\subsubsection{Multiplicity of Features}\label{sec:InputCount}
For most tasks, a model only processes a single input, such as an image, a sentence, or an acoustic sequence. We refer to such a mechanism as \textbf{singular features attention}. Other models are designed to use attention based on multiple inputs to allow one to introduce more information into the model that can be exploited in various ways. However, this does imply the presence of multiple feature matrices that require special attention mechanisms to be fully used. For example, \cite{NIPS2016_6202} introduces a concept named \textbf{co-attention} to allow the proposed visual question answering (VQA) model to jointly attend to both an image and a question.

Co-attention mechanisms can generally be split up into two groups \cite{fan2018multi}: \textbf{coarse-grained co-attention} and \textbf{fine-grained co-attention}. The difference between the two groups is the way attention scores are calculated based on the two feature matrices. Coarse-grained attention mechanisms use a compact representation of one feature matrix as a query when attending to the other feature vectors. Fine-grained co-attention, on the other hand, uses all feature vectors of one input as queries. As such, no information is lost, which is why these mechanisms are called fine-grained.

\begin{figure}
    \centering
    \includegraphics[scale=0.65]{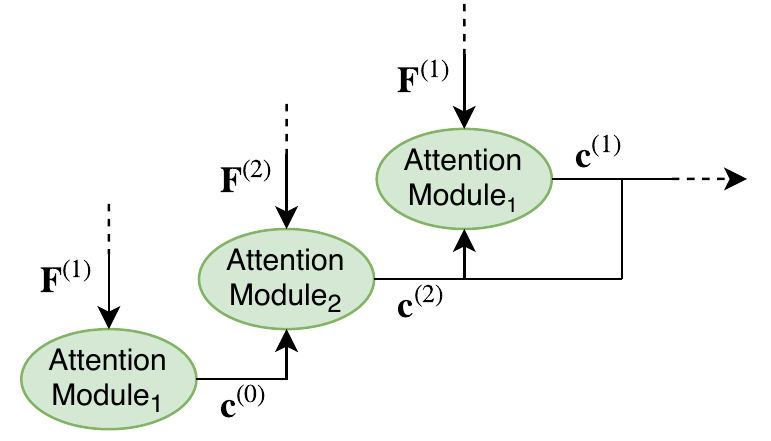}
    \caption{An illustration of alternating co-attention.}
    \label{fig:AlternatingCo-Attention}
\end{figure}

As an example of coarse-grained co-attention, \cite{NIPS2016_6202} proposes an \textbf{alternating co-attention} mechanism that uses the context vector (which is a compact representation) from one attention module as the query for the other module, and vice versa. Alternating co-attention is presented in Fig. \ref{fig:AlternatingCo-Attention}. 
Given a set of two input matrices $\bm X^{(1)}$ and $\bm X^{(2)}$, features are extracted by a feature model to produce the feature matrices $\bm F^{(1)} \in \mathbb{R}^{d_f^{(1)} \times n_f^{(1)}}$ and $\bm F^{(2)} \in \mathbb{R}^{d_f^{(2)} \times n_f^{(2)}}$, where $d_f^{(1)}$ and $d_f^{(2)}$ represent, respectively, the dimension of the feature vectors extracted from the first and second inputs, while $n_f^{(1)}$ and $n_f^{(2)}$ represent, respectively, the amount of feature vectors extracted from the first and second inputs. In \cite{NIPS2016_6202}, co-attention is used for VQA, so the two input matrices are the image data and the question data, for which the feature model for the image consists of a CNN model, and the feature model for the question consists of word embeddings, a convolutional layer, a pooling layer, and an LSTM model. Firstly, attention is calculated for the first set of features $\bm F^{(1)}$ without the use of a query (Attention Module\textsubscript{1} in Fig. \ref{fig:AlternatingCo-Attention}). In \cite{NIPS2016_6202}, an adjusted additive attention score function is used for this attention mechanism. The general form of the regular additive score function can be seen in (\ref{equation:MultiplicityFeaturesAdditive}).
\begin{equation}\label{equation:MultiplicityFeaturesAdditive}
\def\sss{\scriptscriptstyle}
\setstackgap{L}{11pt}
\def\stacktype{L}
    \text{score}(\hspace{-3pt}\stackunder{\bm q}{\sss d_q \times 1}, \stackunder{\bm k_l}{\sss d_k \times 1}\hspace{-3pt}) = \stackunder{\bm w^T}{\sss 1 \times d_w} \hspace{-4pt}\times \text{act}(\hspace{-4pt}\stackunder{\bm W_1}{\sss d_w \times d_{q}} \hspace{-3pt}\times\hspace{-1pt}\stackunder{\bm q}{\sss d_{q} \times 1} + \stackunder{\bm W_2}{\sss d_w \times d_{k}} \hspace{-3pt}\times \stackunder{\bm k_{l}}{\sss d_{k} \times 1}\hspace{-2pt} + \hspace{-4pt}\stackunder{\bm b}{\sss d_{w} \times 1}\hspace{-4pt}),
\end{equation}
where $\text{act}()$ is a non-linear activation function, and $\bm w \in \mathbb{R}^{d_w}$, $\bm W_1 \in \mathbb{R}^{d_w \times d_{q}}$, $\bm W_2 \in \mathbb{R}^{d_w \times d_{k}}$, and $\bm b \in \mathbb{R}^{d_w}$ are trainable weights matrices, for which $d_w$ is a predefined dimension of the weight matrices. A variant of this score function adapted to be calculated without a query for the application at hand can be seen in (\ref{equation:AlternatingCo-AttentionScores1}).
\begin{equation}\label{equation:AlternatingCo-AttentionScores1}
\def\sss{\scriptscriptstyle}
\setstackgap{L}{11pt}
\def\stacktype{L}
    \stackunder{e^{(0)}_l}{\sss 1 \times 1} = \stackunder{\bm w^{(1)T}}{\sss 1 \times d_w} \times \text{act}(\hspace{-1pt}\stackunder{\bm W^{(1)}}{\sss d_w \times d_{k}^{(1)}} \hspace{-3pt}\times \stackunder{\bm k_{l}^{(1)}}{\sss d_{k}^{(1)} \times 1} + \stackunder{\bm b^{(1)}}{\sss d_{w} \times 1}\hspace{-1pt}),
\end{equation}
where $\bm w^{(1)} \in \mathbb{R}^{d_w}$, $\bm W^{(1)} \in \mathbb{R}^{d_w \times d_{k}^{(1)}}$, and $\bm b^{(1)} \in \mathbb{R}^{d_w}$ are trainable weight matrices for Attention Module\textsubscript{1}, $\bm k_{l}^{(1)} \in \mathbb{R}^{d_k^{(1)}}$ is the $l$th column of the keys matrix $\bm K^{(1)}$ that was obtained from $\bm F^{(1)}$ via a linear transformation  (see (\ref{equation:Keys&Values})), for which $d_w$ is a prespecified dimension of the weight matrices and $d_k^{(1)}$ is a prespecified dimension of the key vectors.

Perhaps one may wonder why the query is absent when calculating attention in this manner. Essentially, the query in this attention model is learned alongside the other trainable parameters. As such, the query can be interpreted as a general question: "Which feature vectors contain the most important information?". This is also known as a \textbf{self-attentive mechanism}, since attention is calculated based only on the feature vectors themselves. Self-attention is explained in more detail in Subsection \ref{sec:OutputTypes}.

\begin{figure}
    \centering
    \includegraphics[scale=0.58]{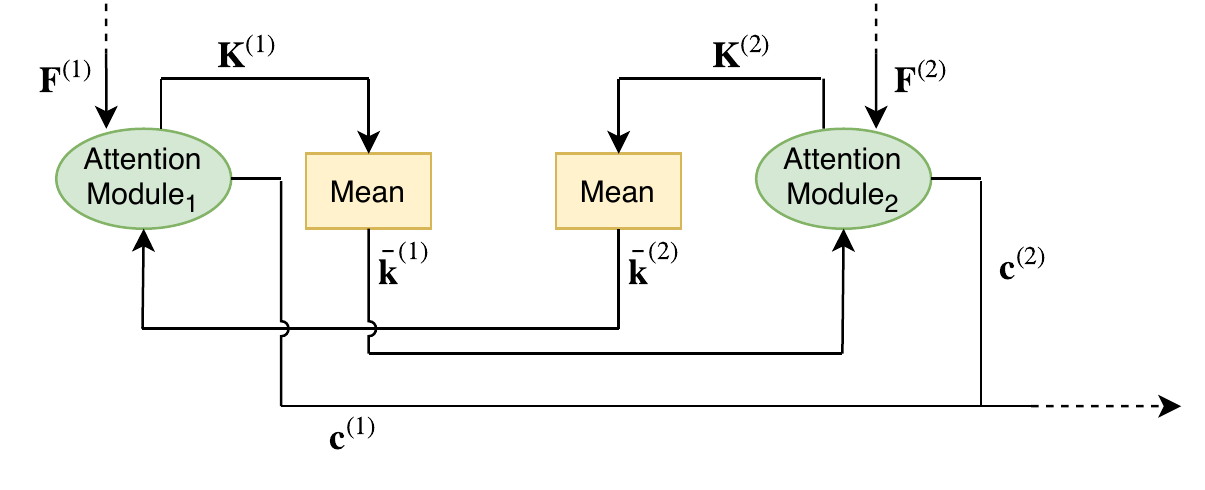}
    \caption{An illustration of interactive co-attention.}
    \label{fig:InteractiveCo-Attention}
\end{figure}

The scores are combined with an alignment function (see (\ref{equation:GeneralAttentionWeight})), such as the softmax function, to create attention weights used to calculate the context vector $\bm c^{(0)} \in \mathbb{R}^{d_v^{(1)}}$ (see (\ref{equation:GeneralContextVector})). This context vector is not used as the output of the attention model, but rather as a query for calculating the context vector $\bm c^{(2)} \in \mathbb{R}^{d_v^{(2)}}$, based on the second feature matrix $\bm F^{(2)}$, where $d_v^{(2)}$ is the dimension of the value vectors obtained from $\bm F^{(2)}$ via a linear transformation (see (\ref{equation:Keys&Values})). For this module (Attention Module\textsubscript{2} in Fig. \ref{fig:AlternatingCo-Attention}), attention scores are calculated using another score function with $\bm c_0$ as query input, as presented in (\ref{equation:AlternatingCo-AttentionScores2}). Any function can be used in this situation, but an additive function is used in \cite{NIPS2016_6202}.
\begin{equation}\label{equation:AlternatingCo-AttentionScores2}
    \def\sss{\scriptscriptstyle}
    \setstackgap{L}{11pt}
    \def\stacktype{L}
   \stackunder{e_{l}^{(2)}}{\sss 1 \times 1} = \text{score}(\stackunder{\bm c^{(0)}}{\sss d_v^{(1)} \times 1}, \stackunder{\bm k_l^{(2)}}{\sss d_k^{(2)} \times 1}).
\end{equation}
These attention scores are then used to calculate attention weights using, for example, a softmax function as alignment function, after which the context vector $\bm c^{(2)}$ can be derived as a weighted average of the second set of value vectors. Finally, the context vector $\bm c^{(2)}$ is used as a query for the first attention module, which will produce the context vector $\bm c^{(1)}$ for the first feature matrix $\bm F^{(1)}$. Attention scores are calculated according to (\ref{equation:AlternatingCo-AttentionScores3}). In \cite{NIPS2016_6202}, the same function and weight matrices as seen in (\ref{equation:AlternatingCo-AttentionScores1}) are used, but with an added query making it the same as the general additive score function (see (\ref{equation:MultiplicityFeaturesAdditive})). The rest of the attention calculation is similar as before.
\begin{equation}\label{equation:AlternatingCo-AttentionScores3}
    \def\sss{\scriptscriptstyle}
    \setstackgap{L}{11pt}
    \def\stacktype{L}
   \stackunder{e_{l}^{(1)}}{\sss 1 \times 1} = \text{score}(\stackunder{\bm c^{(2)}}{\sss d_v^{(2)} \times 1}, \stackunder{\bm k_l^{(1)}}{\sss d_k^{(1)} \times 1}).
\end{equation}
The produced context vectors $\bm c^{(1)}$ and $\bm c^{(2)}$ are concatenated and used for prediction in the output model. Alternating co-attention inherently contains a form of sequentiality due to the fact that context vectors need to be calculated one after another. This may come with a computational disadvantage since it is not possible to parallelize. Instead of using a sequential mechanism like alternating co-attention, \cite{ma2017interactive} proposes the \textbf{interactive co-attention} mechanism that can calculate attention on both feature matrices in parallel, as depicted in Fig. \ref{fig:InteractiveCo-Attention}. Instead of using the context vectors as queries, unweighted averages of the key vectors are used as queries. The calculation of the average keys are provided in (\ref{equation:Co-AttentionKeyAverages}), and the calculation of the attention scores are shown in (\ref{equation:Co-AttentionScoresAlt}). Any score function can be used in this case, but an additive score function is used in \cite{ma2017interactive}.
\begin{equation}\label{equation:Co-AttentionKeyAverages}
\def\sss{\scriptscriptstyle}
\setstackgap{L}{13pt}
\def\stacktype{L}
    \stackunder{\bar{\bm k}^{(1)}}{\sss d_{k}^{(1)} \times 1}= \frac{1}{n^{(1)}_f}\sum^{n^{(1)}_f}_{l=1}\stackunder{\bm k_{l}^{(1)}}{\sss d_{k}^{(1)} \times 1},
    \hspace{20pt}
    \stackunder{\bar{\bm k}^{(2)}}{\sss d_{k}^{(2)} \times 1}= \frac{1}{n^{(2)}_f}\sum^{n^{(2)}_f}_{l=1}\stackunder{\bm k_{l}^{(2)}}{\sss d_{k}^{(2)} \times 1};
\end{equation}
\begin{equation}\label{equation:Co-AttentionScoresAlt}
\def\sss{\scriptscriptstyle}
\setstackgap{L}{13pt}
\def\stacktype{L}
    \stackunder{e^{(1)}_l}{\sss 1 \times 1} = \text{score}(\stackunder{\bar{\bm k}^{(2)}}{\sss d_{k}^{(2)} \times 1}, \stackunder{\bm k_l^{(1)}}{\sss d_k^{(1)} \times 1}),
    \hspace{5pt}
    \stackunder{e^{(2)}_l}{\sss 1 \times 1} = \text{score}(\stackunder{\bar{\bm k}^{(1)}}{\sss d_{k}^{(1)} \times 1}, \stackunder{\bm k_l^{(2)}}{\sss d_k^{(2)} \times 1}).
\end{equation}
From the attention scores, attention weights are created via an alignment function, and are used to produce the context vectors $\bm c^{(1)}$ and $\bm c^{(2)}$.

\begin{figure}
    \centering
    \includegraphics[scale=0.6]{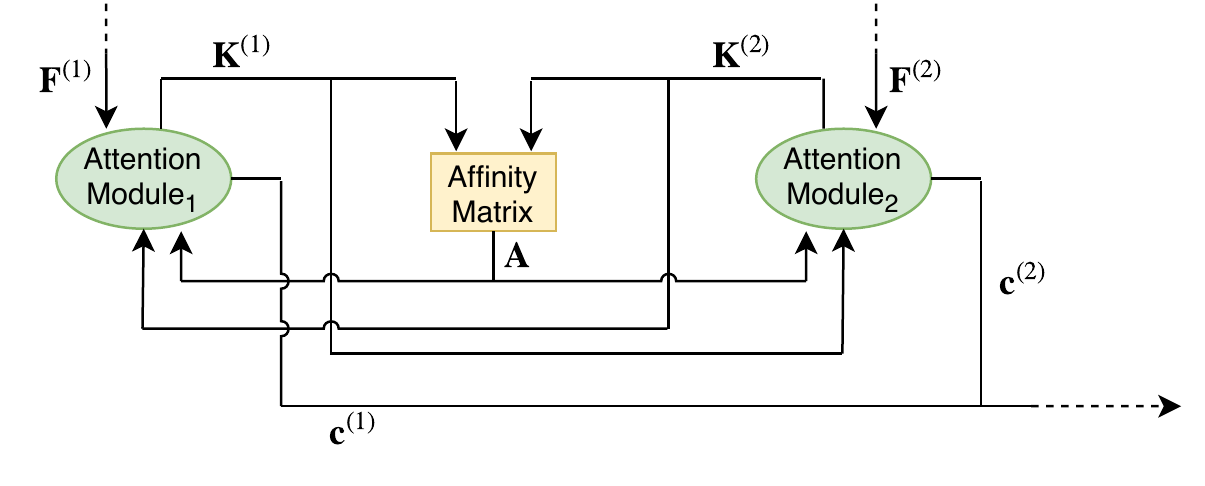}
    \caption{An illustration of parallel co-attention.}
    \label{fig:ParallelCo-Attention}
\end{figure}

While coarse-grained co-attention mechanisms use a compact representation of one input to use as a query when calculating attention for another input, fine-grained co-attention considers every element of each input individually when calculating attention scores. In this case, the query becomes a matrix. An example of fine-grained co-attention is \textbf{parallel co-attention} \cite{NIPS2016_6202}. Similarly to interactive co-attention, parallel co-attention calculates attention on the two feature matrices at the same time, as shown in Fig. \ref{fig:ParallelCo-Attention}. We start by evaluating the keys matrices $\bm K^{(1)} \in \mathbb{R}^{d_{k}^{(1)} \times n_{f}^{(1)}}$ and $\bm K^{(2)} \in \mathbb{R}^{d_{k}^{(2)} \times n_{f}^{(2)}}$ that are obtained by linearly transforming the feature matrices $\bm F^{(1)}$ and $\bm F^{(2)}$, where $d_k^{(1)}$ and $d_k^{(2)}$ are prespecified dimensions of the keys. The idea is to use the keys matrix from one input as the query for calculating attention on the other input. However, since $\bm K^{(1)}$ and $\bm K^{(2)}$ have completely different dimensions, an affinity matrix $\bm A \in \mathbb{R}^{n_{f}^{(1)} \times n_{f}^{(2)}}$ is calculated that is used to essentially translate one keys matrix to the space of the other keys. In \cite{NIPS2016_6202}, $\bm A$ is calculated as shown in (\ref{equation:AffinityMatrix1}).
\begin{equation}\label{equation:AffinityMatrix1}
\def\sss{\scriptscriptstyle}
\setstackgap{L}{11pt}
\def\stacktype{L}
    \stackunder{\bm A}{\sss n_{f}^{(1)} \times n_{f}^{(2)}} = \text{act}(\stackunder{\bm K^{{(1)}^T}}{\sss n_{f}^{(1)} \times d_{k}^{(1)}} \times \stackunder{\bm W_A}{\sss d_{k}^{(1)} \times d_{k}^{(2)}} \times \stackunder{\bm K^{(2)}}{\sss d_{k}^{(2)} \times n_{f}^{(2)}}),
\end{equation}
where $\bm W_A \in \mathbb{R}^{d_k^{(1)} \times d_k^{(2)}}$ is a trainable weights matrix and $\text{act}()$ is an activation function for which the $\text{tanh}()$ function is used in \cite{NIPS2016_6202}. \cite{seo2016bidirectional} proposes a different way of calculating this matrix, i.e., one can use (\ref{equation:AffinityMatrix2}) to calculate each individual element $A_{i,j}$ of the matrix $\bm A$.
\begin{equation}\label{equation:AffinityMatrix2}
\def\sss{\scriptscriptstyle}
\setstackgap{L}{11pt}
\def\stacktype{L}
    \stackunder{A_{i,j}}{\sss 1 \times 1} = \stackunder{\bm w_A^T}{\sss 1 \times 3 d_{k}} \times \text{concat}(\stackunder{\bm k_{i}^{(1)}}{\sss d_k \times 1}, \stackunder{\bm k_{j}^{(2)}}{\sss d_k \times 1}, \stackunder{\bm k_{i}^{(1)}}{\sss d_k \times 1} \circ \stackunder{\bm k_{j}^{(2)}}{\sss d_k \times 1} ),
\end{equation}
where $\bm w_A \in \mathbb{R}^{3 d_{k}}$ denotes a trainable vector of weights, $\text{concat}()$ denotes vector concatenation, and $\circ$ denotes element-wise multiplication, also known as the Hadamard product. Note that the keys of each keys matrix in this case must have the same dimension $d_k$ for the element-wise multiplication to work. The affinity matrix can be interpreted as a similarity matrix for the columns of the two keys matrices, and helps translate, for example, image keys to the same space as the keys of the words in a sentence, and vice versa. The vectors of attention scores $\bm e^{(1)}$ and $\bm e^{(2)}$ can be calculated using an altered version of the additive score function as presented in (\ref{equation:Co-AttentionScores1}) and (\ref{equation:Co-AttentionScores2}). The previous attention score examples in this survey all used a score function to calculate each attention score for each value vector individually. However, (\ref{equation:Co-AttentionScores1}) and (\ref{equation:Co-AttentionScores2}) are used to calculate the complete vector of all attention scores. Essentially, the attention scores are calculated in an aggregated form.
\begin{equation}\label{equation:Co-AttentionScores1}
\def\sss{\scriptscriptstyle}
\setstackgap{L}{11pt}
\def\stacktype{L}
    \stackunder{\bm e^{(1)}}{\sss 1 \times n_{f}^{(1)}} = \stackunder{\bm w_{1}}{\sss 1 \times d_w} \hspace{-4pt}\times \text{act}(\hspace{-5pt}\stackunder{\bm W_2}{\sss d_w \times d_{k}^{(2)}} \hspace{-4pt}\times\hspace{-4pt} \stackunder{\bm K^{(2)}}{\sss d_{k}^{(2)} \times n_{f}^{(2)}} \hspace{-4pt}\times\hspace{-4pt} \stackunder{\bm A^T}{\sss n_{f}^{(2)} \times n_{f}^{(1)}} \hspace{-4pt}+\hspace{-4pt}\stackunder{\bm W_1}{\sss d_w \times d_{k}^{(1)}} \hspace{-4pt}\times\hspace{-4pt} \stackunder{\bm K^{(1)}}{\sss d_{k}^{(1)} \times n_{f}^{(1)}} \hspace{-4pt});
\end{equation}
\begin{equation}\label{equation:Co-AttentionScores2}
\def\sss{\scriptscriptstyle}
\setstackgap{L}{11pt}
\def\stacktype{L}
    \stackunder{\bm e^{(2)}}{\sss 1 \times n_{f}^{(2)}} = \stackunder{\bm w_{2}}{\sss 1 \times d_w} \hspace{-4pt}\times \text{act}(\hspace{-5pt}\stackunder{\bm W_1}{\sss d_w \times d_{k}^{(1)}} \hspace{-4pt}\times\hspace{-4pt} \stackunder{\bm K^{(1)}}{\sss d_{k}^{(1)} \times n_{f}^{(1)}} \hspace{-4pt}\times\hspace{-4pt}  \stackunder{\bm A}{\sss n_{f}^{(1)} \times n_{f}^{(2)}}\hspace{-4pt}+\hspace{-4pt}\stackunder{\bm W_2}{\sss d_w \times d_{k}^{(2)}} \hspace{-4pt}\times\hspace{-4pt} \stackunder{\bm K^{(2)}}{\sss d_{k}^{(2)} \times n_{f}^{(2)}} \hspace{-5pt}),
\end{equation}
where $\bm w_1 \in \mathbb{R}^{d_w}$, $\bm w_2 \in \mathbb{R}^{d_w}$, $\bm W_1 \in \mathbb{R}^{d_w \times d_{k}^{(1)}}$, and $\bm W_2 \in \mathbb{R}^{d_w \times d_{k}^{(2)}}$ are trainable weight matrices, for which $d_w$ is a prespecified dimension of the weight matrices. Note that $\text{tanh}()$ is used in \cite{NIPS2016_6202} for the activation function, and the feature matrices are used as the key matrices. In that case, the affinity matrix $\bm A$ can be seen as a translator between feature spaces. As mentioned before, the affinity matrix is essentially a similarity matrix for the key vectors of the two inputs. In \cite{fan2018multi}, this fact is used to propose a different way of determining attention scores. Namely, one could take the maximum similarity value in a row or column as the attention score, as shown in (\ref{equation:Co-AttentionScores3}).
\begin{equation}\label{equation:Co-AttentionScores3}
\def\sss{\scriptscriptstyle}
\setstackgap{L}{8pt}
\def\stacktype{L}
    \stackunder{e_i^{(1)}}{\sss 1 \times 1} = \stackunder{\text{max}}{\sss j=1, \dots ,n_f^{(2)}} \stackunder{A_{i,j}}{\sss 1 \times 1},
    \hspace{20pt}
    \stackunder{e_j^{(2)}}{\sss 1 \times 1} = \stackunder{\text{max}}{\sss i=1, \dots ,n_f^{(1)}} \stackunder{A_{i,j}}{\sss 1 \times 1}.
\end{equation}

Next, the attention scores are used to calculate attention weights using an alignment function, so that two context vectors $\bm c^{(1)}$ and $\bm c^{(2)}$ can be derived as weighted averages of the value vectors that are obtained from linearly transforming the features. For the alignment function, \cite{NIPS2016_6202} proposes to use a softmax function, and the value vectors are simply set equal to the feature vectors. The resulting context vectors can be either concatenated or added together.

Finally, coarse-grained and fine-grained co-attention can be combined to create an even more complex co-attention mechanism. \cite{fan2018multi} proposes the \textbf{multi-grained co-attention} mechanism that calculates both coarse-grained and fine-grained co-attention for two inputs. Each mechanism produces one context vector per input. The four resulting context vectors are concatenated and used in the output model for prediction.

A mechanism separate from co-attention that still uses multiple inputs is the \textbf{rotatory attention} mechanism \cite{zheng2018left}. This technique is typically used in a text sentiment analysis setting where there are three inputs involved: the phrase for which the sentiment needs to be determined (target phrase), the text before the target phrase (left context), and the text after the target phrase (right context). The words in these three inputs are all encoded by the feature model, producing the following feature matrices: $\bm F^t = [\bm f^t_1, \dots ,\bm f^t_{n^t_f}] \in \mathbb{R}^{d_f^t \times n_f^t}$, $\bm F^l= [\bm f^l_1, \dots ,\bm f^l_{n^l_f}] \in \mathbb{R}^{d_f^l \times n_f^l}$, and $\bm F^r= [\bm f^r_1, \dots ,\bm f^r_{n^r_f}] \in \mathbb{R}^{d_f^r \times n_f^r}$, for the target phrase words, left context words, and right context words, respectively, where $d_f^t$, $d_f^l$, and $d_f^r$ represent the dimensions of the feature vectors for the corresponding inputs, and $n_f^t$, $n_f^l$, and $n_f^r$ represent the number of feature vectors for the corresponding inputs. The feature model used in \cite{zheng2018left} consists of word embeddings and separate Bi-LSTM models for the target phrase, the left context, and the right context. This means that the feature vectors are in fact the hidden state vectors obtained from the Bi-LSTM models. Using these features, the idea is to extract a single vector $\bm r$ from the inputs such that a softmax layer can be used for classification. As such, we are now faced with two challenges: how to represent the inputs as a single vector, and how to incorporate the information from the left and right context into that vector. \cite{zheng2018left} proposes to use the rotatory attention mechanism for this purpose.

Firstly, a single target phrase representation is created by using a pooling layer that takes the average over the columns of $\bm F^t$, as shown in (\ref{equation:TargetAverage}).
\begin{equation}\label{equation:TargetAverage}
\def\sss{\scriptscriptstyle}
\setstackgap{L}{11pt}
\def\stacktype{L}
    \stackunder{\bm r^t}{\sss d_{f}^{t} \times 1}= \frac{1}{n^t_f}\sum^{n^t_f}_{i=1}\stackunder{\bm f^t_{i}}{\sss d_{f}^{t} \times 1}.
\end{equation}
$\bm r^t$ is then used as a query to create a context vector out of the left and right contexts, separately. For example, for the left context, the key vectors $\bm k_1^l , \dots , \bm k_{n_f^l}^l \in \mathbb{R}^{d_k^l}$ and value vectors $\bm v_1^l , \dots , \bm v_{n_f^l}^l \in \mathbb{R}^{d_v^l}$ are extracted from the left context feature vectors $\bm f_1^l , \dots , \bm f_{n_f^l}^l \in \mathbb{R}^{d_f^l}$, similarly as before, where $d_k^l$ and $d_v^l$ are the dimensions of the key and value vectors, respectively. Note that \cite{zheng2018left} proposes to use the original feature vectors as keys and values, meaning that the linear transformation consists of a multiplication by an identity matrix. Next, the scores are calculated using (\ref{equation:rotatoryAttentionScore}).
\begin{equation}\label{equation:rotatoryAttentionScore}
    \def\sss{\scriptscriptstyle}
    \setstackgap{L}{8pt}
    \def\stacktype{L}
   \stackunder{e_{i}^l}{\sss 1 \times 1} = \text{score}(\stackunder{\bm r^t}{\sss d_{f}^{t} \times 1}, \stackunder{\bm k_i^l}{\sss d_k^l \times 1}).
\end{equation}
For the score function, \cite{zheng2018left} proposes to use an activated general score function \cite{ma2017interactive} with a tanh activation function. The attention scores can be combined with an alignment function and the corresponding value vectors to produce the context vector $\bm r^l \in \mathbb{R}^{d_v^l}$. The alignment function used in \cite{zheng2018left} takes the form of a softmax function.  An analogous procedure can be performed to obtain the representation of the right context, $\bm r^r$. These two context representations can then be used to create new representations of the target phrase, again, using attention. Firstly, the key vectors $\bm k_1^t , \dots , \bm k_{n_f^t}^t \in \mathbb{R}^{d_k^t}$ and value vectors $\bm v_1^t , \dots , \bm v_{n_f^t}^t \in \mathbb{R}^{d_v^t}$ are extracted from the target phrase feature vectors $\bm f_1^t , \dots , \bm f_{n_f^t}^t \in \mathbb{R}^{d_f^t}$, similarly as before, using a linear transformation, where $d_k^t$ and $d_v^t$ are the dimensions of the key and value vectors, respectively. Note, again, that the original feature vectors as keys and values in \cite{zheng2018left}. The attention scores for the left-aware target representation are then calculated using (\ref{equation:rotatoryAwareAttentionScore}).
\begin{equation}\label{equation:rotatoryAwareAttentionScore}
    \def\sss{\scriptscriptstyle}
    \setstackgap{L}{8pt}
    \def\stacktype{L}
   \stackunder{e_{i}^{l_t}}{\sss 1 \times 1} = \text{score}(\stackunder{\bm r^l}{\sss d_{v}^{l} \times 1}, \stackunder{\bm k_i^t}{\sss d_k^t \times 1}).
\end{equation}
The attention scores can be combined with an alignment function and the corresponding value vectors to produce the context vector $\bm r^{l_t} \in \mathbb{R}^{d_v^t}$. For this attention calculation, \cite{ma2017interactive} proposes to use the same score and alignment functions as before. The right-aware target representation $\bm r^{r_t}$ can be calculated in a similar manner. Finally, to obtain the full representation vector $\bm r$ that is used to determine the classification, the vectors $\bm r^{l}$, $\bm r^{r}$, $\bm r^{l_t}$, and $\bm r^{r_t}$ are concatenated together, as shown in (\ref{equation:rotatoryFinal}).
\begin{equation}\label{equation:rotatoryFinal}
    \def\sss{\scriptscriptstyle}
    \setstackgap{L}{8pt}
    \def\stacktype{L}
    \stackunder{\bm r}{\sss (d_v^l + d_v^r + d_v^t + d_v^t) \times 1} = \text{concat}(\stackunder{\bm r^l}{\sss d_v^l \times 1}, \stackunder{\bm r^r}{\sss d_v^r \times 1}, \stackunder{\bm r^{l_t}}{\sss d_v^t \times 1}, \stackunder{\bm r^{r_t}}{\sss d_v^t \times 1}).
\end{equation}

To summarize, rotatory attention uses the target phrase to compute new representations for the left and right context using attention, and then uses these left and right representations to calculate new representations for the target phrase. The first step is designed to capture the words in the left and right contexts that are most important to the target phrase. The second step is there to capture the most important information in the actual target phrase itself. Essentially, the mechanism rotates attention between the target and the contexts to improve the representations.

There are many applications where combining information from different inputs into a single model can be highly beneficial. For example, in the field of medical data, there are often many different types of data available, such as various scans or documents, that can provide different types of information. In \cite{jing-etal-2018-automatic}, a co-attention mechanism is used for automatic medical report generation to attend to both images and semantic tags simultaneously. Similarly, in \cite{gao2019camp}, a co-attention model is proposed that combines general demographics features and patient medical history features to predict future health information. Additionally, an ablation study is used in \cite{gao2019camp} to show that the co-attention part of the model specifically improves performance. A field where multi-feature attention has been extensively explored is the domain of recommender systems. For example, in \cite{10.1145/3219819.3220086}, a co-attention network is proposed that attends to both product reviews and the reviews a user has written. In \cite{10.1145/3308558.3313513}, a model is proposed for video recommendation that attends to both user features and video features. Co-attention techniques have also been used in combination with graph networks for the purpose of, for example, reading comprehension across multiple documents \cite{tu-etal-2019-multi} and fake news detection \cite{lu2020gcan}. In comparison to co-attention, rotatory attention has typically been explored only in the field of sentiment analysis, which is most likely due to the specific structure of the data that is necessary to use this technique. An implementation of rotatory attention is proposed in \cite{wallaart2019hybrid} for sentiment analysis, where the mechanism is extended by repeating the attention rotation to iteratively further improve the representations.

\subsubsection{Feature Levels}\label{sec:InputHierarchical}
The previously discussed attention mechanisms process data at a single level. We refer to these attention techniques as \textbf{single-level attention} mechanisms. However, some data types can be analyzed and represented on multiple levels. For example, when analyzing documents, one can analyze the document at the sentence level, word level, or even the character level. When representations or embeddings of all these levels are available, one can exploit the extra levels of information. For example, one could choose to perform translation based on either just the characters, or just the words of the sentence. However, in \cite{zhao2018attention}, a technique named \textbf{attention-via-attention} is introduced that allows one to incorporate information from both the character, and the word levels. The idea is to predict the sentence translation character-by-character, while also incorporating information from a word-level attention module.
\begin{figure}
    \centering
    \includegraphics[scale=0.64]{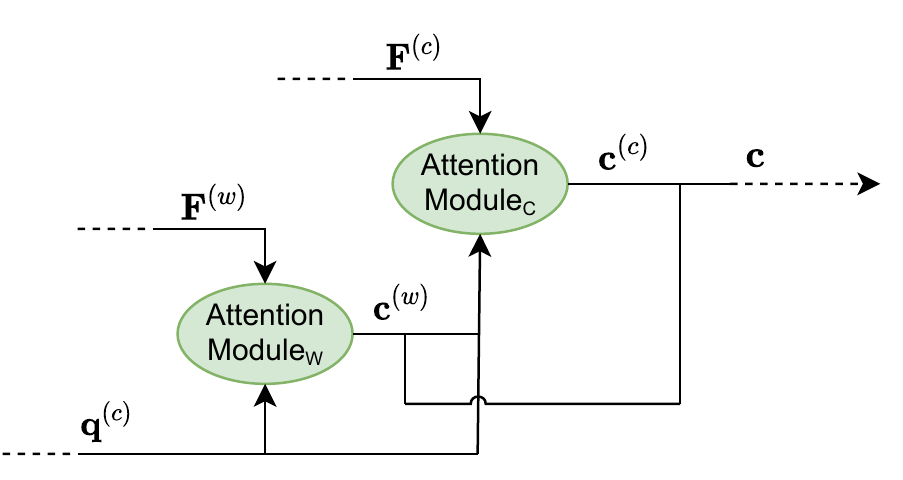}
    \caption{An illustration of attention-via-attention.}
    \label{fig:AttentionViaAttention}
\end{figure}

To begin with, a feature model (consisting of, for example, word embeddings and RNNs) is used to encode the input sentence into both a character-level feature matrix $\bm F^{(c)} \in \mathbb{R}^{d_{f}^{(c)} \times n_f^{(c)}}$, and a word-level feature matrix $\bm F^{(w)} \in \mathbb{R}^{d_{f}^{(w)} \times n_f^{(w)}}$, where $d_{f}^{(c)}$ and $n_f^{(c)}$ represent, respectively, the dimension of the embeddings of the characters, and the number of characters, while $d_{f}^{(w)}$ and $n_f^{(w)}$ represent the same but at the word level. It is crucial for this method that each level in the data can be represented or embedded. When attempting to predict a character in the translated sentence, a query $\bm q^{(c)} \in \mathbb{R}^{d_{q}}$ is created by the query model (like a character-level RNN), where $d_{q}$ is the dimension of the query vectors. As illustrated in Fig. \ref{fig:AttentionViaAttention}, the query is used to calculate attention on the word-level feature vectors $\bm F^{(w)}$. This generates the context vector $\bm c^{(w)} \in \mathbb{R}^{d_{v}^{(w)}}$, where $d_{v}^{(w)}$ represents the dimension of the value vectors for the word-level attention module. This context vector summarizes which words contain the most important information for predicting the next character. If we know which words are most important, then it becomes easier to identify which characters in the input sentence are most important. Thus, the next step is to attend to the character-level features in $\bm F^{(c)}$, with an additional query input: the word-level context vector $\bm c^{(w)}$. The actual query input for the attention model will therefore be the concatenation of the query $\bm q^{(c)}$ and the word context vector $\bm c^{(w)}$. The output of this character-level attention module is the context vector $\bm c^{(c)}$. The complete context output of the attention model is the concatenation of the word-level, and character-level context vectors.

The attention-via-attention technique uses representations for each level. However, accurate representations may not always be available for each level of the data, or it may be desirable to let the model create the representations during the process by building them from lower level representations. A technique referred to as \textbf{hierarchical attention} \cite{yang-etal-2016-hierarchical} can be used in this situation. Hierarchical attention is another technique that allows one to apply attention on different levels of the data. Yet, the exact mechanisms work quite differently compared to attention-via-attention. The idea is to start at the lowest level, and then create representations, or summaries, of the next level using attention. This process is repeated till the highest level is reached. To make this a little clearer, suppose one attempts to create a model for document classification, similarly to the implementation from \cite{yang-etal-2016-hierarchical}. We analyze a document containing $n_S$ sentences, with the $s$th sentence containing $n_s$ words, for $s=1,\dots, n_S$. One could use attention based on just the collection of words to classify the document. However, a significant amount of important context is then left out of the analysis, since the model will consider all words as a single long sentence, and will therefore not consider the context within the separate sentences. Instead, one can use the hierarchical structure of a document (words form sentences, and sentences form the document).

\begin{figure}
    \centering
    \includegraphics[scale=0.6]{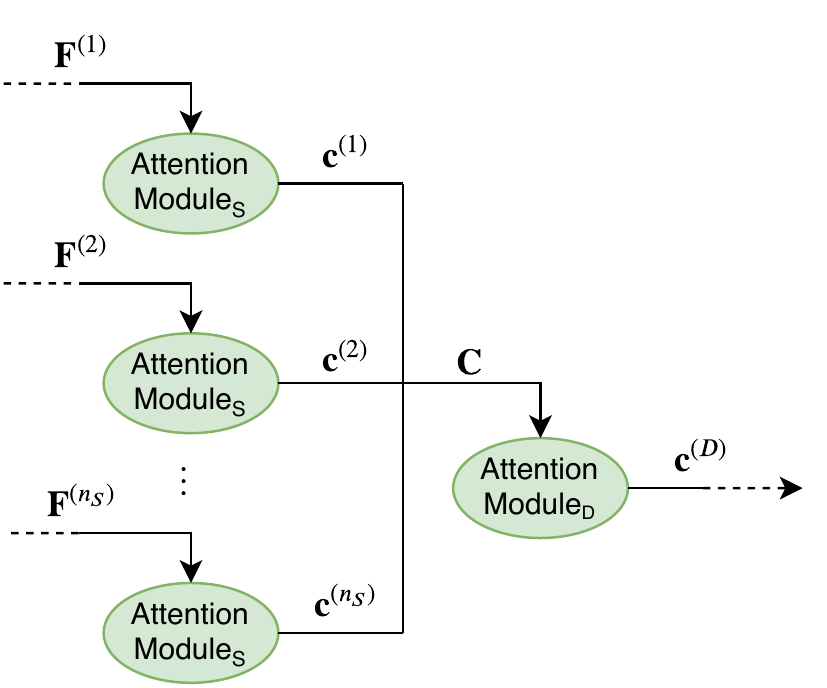}
    \caption{An illustration of hierarchical attention.}
    \label{fig:HierarchicalAttention}
\end{figure}

Fig. \ref{fig:HierarchicalAttention} illustrates the structure of hierarchical attention. For each sentence in the document, a sentence representation $\bm c^{(s)} \in \mathbb{R}^{d_{v}^{(S)}}$ is produced, for $s = 1, \dots, n_S$, where $d_{v}^{(S)}$ is the dimension of the value vectors used in the attention model for sentence representations (Attention Module\textsubscript{S} in Fig. \ref{fig:HierarchicalAttention}). The representation is a context vector from an attention module that essentially summarizes the sentence. Each sentence is first put through a feature model to extract the feature matrix $\bm F^{(s)} \in \mathbb{R}^{d_{f}^{(S)} \times n_{s}}$, for $s = 1, \dots, n_S$, where $d_{f}^{(S)}$ represents the dimension of the feature vector for each word, and $n_{s}$ represents the amount of words in sentence $s$. For extra clarification, the columns of $\bm F^{(s)}$ are feature vectors that correspond to the words in sentence $s$. As shown in Fig. \ref{fig:HierarchicalAttention}, each feature matrix $\bm F^{(s)}$ is used as input for an attention model, which produces the context vector $\bm c^{(s)}$, for each $s = 1, \dots, n_S$. No queries are used in this step, so it can be considered a self-attentive mechanism. The context vectors are essentially summaries of the words in the sentences. The matrix of context vectors $\bm C = [\bm c^{(1)}, \dots, \bm c^{(n_S)}] \in \mathbb{R}^{d_{v}^{(S)} \times n_{S}}$ is constructed by grouping all the obtained context vectors together as columns. Finally, attention is calculated using $\bm C$ as feature input, producing the representation of the entire document in the context vector $\bm c^{(D)} \in \mathbb{R}^{d_{v}^{(D)}}$, where $d_{v}^{(D)}$ is the dimension of the value vectors in the attention model for document representation (Attention Module\textsubscript{D} in Fig. \ref{fig:HierarchicalAttention}). This context vector can be used to classify the document, since it is essentially a summary of all the sentences (and therefore also the words) in the document.

Multi-level models can be used in a variety of tasks. For example, in \cite{ying2018sequential}, hierarchical attention is used in a recommender system to model user preferences at the long-term level and the short-term level. Similarly, \cite{8700213} proposes a hierarchical model for recommending social media images based on user preferences. Hierarchical attention has also been successfully applied in other domains. For example, \cite{wang2016hierarchical} proposes to use hierarchical attention in a video action recognition model to capture motion information at the the long-term level and the short-term level. Furthermore, \cite{li2018hierarchical} proposes a hierarchical attention model for cross-domain sentiment classification. In \cite{xing2018hierarchical}, a hierarchical attention model for chatbot response generation is proposed. Lastly, using image data, \cite{sindagi2019ha} proposes a hierarchical attention model for crowd counting.

\subsubsection{Feature Representations}\label{sec:InputRepresentations}

In a basic attention model, a single embedding or representation model is used to produce feature representations for the model to attend to. This is referred to as \textbf{single-representational attention}. Yet, one may also opt to incorporate multiple representations into the model. In \cite{kiela2018dynamic}, it is argued that allowing a model access to multiple embeddings can allow one to create even higher quality representations. Similarly, \cite{maharjan2018genre} incorporates multiple representations of the same book (textual, syntactic, semantic, visual etc.) into the feature model. Feature representations are an important part of the attention model, but attention can also be an important part of the feature model. The idea is to create a new representation by taking a weighted average of multiple representations, where the weights are determined via attention. This technique is referred to as \textbf{multi-representational attention}, and allows one to create so-called meta-embeddings. Suppose one wants to create a meta-embedding for a word $\bm x$ for which $E$ embeddings $\bm x^{(e_1)}, \dots, \bm x^{(e_E)}$ are available. Each embedding $\bm x^{(e_i)}$ is of size $d_{e_i}$, for $i=1, \dots , E$. Since not all embeddings are of the same size, a transformation is performed to normalize the embedding dimensions. Using embedding-specific weight parameters, each embedding $\bm x^{(e_i)}$ is transformed into the size-normalized embedding $\bm x^{(t_i)} \in \mathbb{R}^{d_t}$, where $d_t$ is the size of every transformed word embedding, as shown in (\ref{equation:MultiRepresentationalTransformation}).
\begin{equation}\label{equation:MultiRepresentationalTransformation}
    \def\sss{\scriptscriptstyle}
    \setstackgap{L}{8pt}
    \def\stacktype{L}
   \stackunder{\bm x^{(t_i)}}{\sss d_t \times 1} = \stackunder{\bm W_{e_i}}{\sss d_t \times d_{e_i}} \times \stackunder{\bm x^{(e_i)}}{\sss d_{e_i} \times 1} + \stackunder{\bm b_{e_i}}{\sss d_t \times 1},
\end{equation}
where $\bm W_{e_i} \in \mathbb{R}^{d_t \times d_{e_i}}$, and $\bm b_{e_i} \in \mathbb{R}^{d_t}$ are trainable, embedding-specific weights matrices. The final embedding $\bm x^{(e)} \in \mathbb{R}^{d_t}$ is a weighted average of the previously calculated transformed representations, as shown in (\ref{equation:MultiRepresentationalAverage}).
\begin{equation}\label{equation:MultiRepresentationalAverage}
    \def\sss{\scriptscriptstyle}
    \setstackgap{L}{8pt}
    \def\stacktype{L}
   \stackunder{\bm x^{(e)}}{\sss d_t \times 1} = \sum_{i=1}^{E} \stackunder{a_i}{\sss 1 \times 1} \times \stackunder{\bm x^{(t_i)}}{\sss d_t \times 1}.
\end{equation}
The final representation $\bm x^{(e)}$ can be interpreted as the context vector from an attention model, meaning that the weights $a_1, \dots, a_E \in \mathbb{R}^{1}$ are attention weights. Attention can be calculated as normally, where the columns of the features matrix $\bm F$ are the transformed representations $\bm x^{(t_1)}, \dots, \bm x^{(t_E)}$. The query in this case can be ignored since it is constant in all cases. Essentially, the query is ``Which representations are the most important?" in every situation. As such, this is a self-attentive mechanism.

While an interesting idea, applications of multi-representational attention are limited. One example of the application of this technique is found in \cite{winata2019learning}, where a multi-representational attention mechanism has been applied to generate multi-lingual meta-embeddings. Another example is \cite{jin2019multi}, where a multi-representational text classification model is proposed that incorporates different representations of the same text. For example, the proposed model uses embeddings from part-of-speech tagging, named entity recognizers, and character-level and word-level embeddings.

\subsection{General Attention Mechanisms}\label{sec:GeneralAttentionMechanisms}
This major category consists of attention mechanisms that can be applied in any type of attention model. The structure of this component can be broken down into the following sub-aspects: the attention score function, the attention alignment, and attention dimensionality.

\begin{table}
\caption{Overview of score function ($\text{score}(\bm q,\bm k_l)$) forms.
}
\begin{center}
\resizebox{\linewidth}{!}{%
\begin{tabular}{|p{90pt}lp{55pt}|}
\hline
Name & Function & Parameters\\ \hline
Additive \newline (Concatenate) \cite{bahdanau2014neural}  & $\bm w^T \times \text{act}(\bm W_1 \times \bm q + \bm W_2 \times \bm k_l) + \bm b )$ & $\bm w \in \mathbb{R}^{d_w} \newline \bm W_1 \in \mathbb{R}^{d_w \times d_q} \newline \bm W_2 \in \mathbb{R}^{d_w \times d_k} \newline \bm b \in \mathbb{R}^{d_w}$ \\ \hline
Multiplicative \newline (Dot-Product) \cite{luong2015effective}  & $\bm q^T \times \bm k_l$ & - \\ \hline
Scaled Multiplicative \cite{vaswani2017attention} & $\frac{\bm q^T \times \bm k_l}{\sqrt{d_k}}$ & - \\ \hline
General \cite{luong2015effective} & $\bm k_l^T \times \bm W \times \bm q$ & $\bm W \in \mathbb{R}^{d_k \times d_q}$ \\ \hline
Biased General \cite{aless2016iterative} & $\bm k_l^T \times  (\bm W \times \bm q + \bm b)$ & $\bm W \in \mathbb{R}^{d_k \times d_q}  \newline \bm b \in \mathbb{R}^{d_k}$ \\ \hline
Activated General \cite{ma2017interactive} & $\text{act}(\bm k_l^T \times \bm W \times \bm q + b)$ & $\bm W \in \mathbb{R}^{d_k \times d_q}, \newline b \in \mathbb{R}^{1}$ \\ \hline
Similarity \cite{graves2014neural} & $\text{similarity}(\bm q, \bm k_l) $ & - \\ \hline

\end{tabular}
}
\end{center}
\label{Table:ScoreFunctions}
\end{table}

\subsubsection{Attention Scoring}\label{sec:ScoreFunction}
The attention score function is a crucial component in how attention is calculated. Various approaches have been developed that each have their own advantages and disadvantages. An overview of these functions is provided in Table \ref{Table:ScoreFunctions}. Each row of Table \ref{Table:ScoreFunctions} presents a possible form for the function $\text{score}(\bm q, \bm k_l)$, as seen in (\ref{equation:AttentionScoreSection}), where $\bm q$ is the query vector, and $\bm k_l$ is the $l$th column of $\bm K$. Note that the score functions presented in this section can be more efficiently calculated in matrix form using $\bm K$ instead of each column separately. Nevertheless, the score functions are presented using $\bm k_l$ to more clearly illustrate the relation between a key and query.
\begin{equation}\label{equation:AttentionScoreSection}
    \def\sss{\scriptscriptstyle}
    \setstackgap{L}{8pt}
    \def\stacktype{L}
   \stackunder{e_{l}}{\sss 1 \times 1} = \text{score}(\stackunder{\bm q}{\sss d_q \times 1}, \stackunder{\bm k_l}{\sss d_k \times 1}).
\end{equation}

Due to their simplicity, the most popular choices for the score function are the \textbf{concatenate score} function \cite{bahdanau2014neural} and the \textbf{multiplicative score} function \cite{luong2015effective}. The multiplicative score function has the advantage of being computationally inexpensive due to highly optimized vector operations. However, the multiplicative function may produce non-optimal results when the dimension $d_k$ is too large \cite{DBLP:journals/corr/BritzGLL17}. When $d_k$ is large, the dot-product between $\bm q$ and $\bm k_l$ can grow large in magnitude. To illustrate this, in \cite{vaswani2017attention}, an example is used where the elements of $\bm q$ and $\bm k_l$ are all normally distributed with a mean equal to zero, and a variance equal to one. Then, the dot-product of the vectors has a variance of $d_k$. A higher variance means a higher chance of numbers that are large in magnitude. When the softmax function of the alignment step is then applied using these large numbers, the gradient will become very small, meaning the model will have trouble converging \cite{vaswani2017attention}. To adjust for this, \cite{vaswani2017attention} proposes to scale the multiplicative function by the factor $\frac{1}{\sqrt{d_k}}$, producing the \textbf{scaled multiplicative score} function.

In \cite{luong2015effective}, the multiplicative score function is extended by introducing a weights matrix $\bm W$. This form, referred to as the \textbf{general score} function, allows for an extra transformation of $\bm k_l$. The \textbf{biased general score} function \cite{aless2016iterative} is a further extension of the general function that introduces a bias weight vector $\bm b$. A final extension on this function named the \textbf{activated general score} function is introduced in \cite{ma2017interactive}, and includes the use of both a bias weight $b$, and an activation function $\text{act}()$.

The previously presented score functions are all based on determining a type of similarity between the key vector and the query vector. As such, more typical similarity measures, such as the Euclidean (L$_2$) distance and cosine similarity, can also be implemented \cite{graves2014neural}. These scoring methods are summarized under the \textbf{similarity score} function which is represented by the $\text{similarity}()$ function.

There typically is no common usage across domains regarding score functions. The choice of score function for a particular task is most often based on empirical experiments. However, there are exceptions when, for example, efficiency is vital. In models where this is the case, the multiplicative or scaled multiplicative score functions are typically the best choice. An example of this is the Transformer model, which is generally computationally expensive.

\subsubsection{Attention Alignment}\label{sec:Alignment}
The attention alignment is the step after the attention scoring. This alignment process directly determines which parts of the input data the model will attend to. The alignment function is denoted as $\text{align}()$ and has various forms. The $\text{align}()$ function takes as input the previously calculated attention score vector $\bm e$ and calculates for each element $e_{l}$ of $\bm e$ the attention weight $a_{l}$. These attention weights can then be used to create the context vector $\bm c$ by taking a weighted average of the value vectors $\bm v_1, \dots, \bm v_{n_f}$:
\begin{equation}\label{equation:AlignContextVector}
    \def\sss{\scriptscriptstyle}
    \setstackgap{L}{8pt}
    \def\stacktype{L}
    \stackunder{\bm c}{\sss d_v \times 1} = \sum^{n_f}_{l=1} \stackunder{a_{l}}{\sss 1 \times 1} \times \stackunder{\bm v_l}{\sss d_v \times 1}.
\end{equation}
The most popular alignment method to calculate these weights is a simple softmax function, as depicted in (\ref{equation:GlobalAttentionWeight}). 
\begin{equation}\label{equation:GlobalAttentionWeight}
    \def\sss{\scriptscriptstyle}
    \setstackgap{L}{8pt}
    \def\stacktype{L}
    \stackunder{a_{l}}{\sss 1 \times 1} = \text{align}(\stackunder{e_{l}}{\sss 1 \times 1}; \stackunder{\bm e}{\sss n_f \times 1}) = \frac{\text{exp}(e_{l})}{\sum^{n_f}_{j=1}\text{exp}(e_{j})}.
\end{equation}
This alignment method is often referred to as \textbf{soft alignment} in computer vision settings \cite{xu2015show}, or \textbf{global alignment} for sequence data \cite{luong2015effective}. Nevertheless, both these terms represent the same function and can be interpreted similarly. Soft/global alignment can be interpreted as the model attending to all feature vectors. For example, the model attends to all regions in an image, or all words in a sentence. Even though the attention model generally does focus more on specific parts of the input, every part of the input will receive at least some amount of attention due to the nature of the softmax function. Furthermore, an advantage of the softmax function is that it introduces a probabilistic interpretation to the input vectors. This allows one to easily analyze which parts of the input are important to the output predictions. 

In contrast to soft/global alignment, other methods aim to achieve a more focused form of alignment. For example, \textbf{hard alignment} \cite{xu2015show}, also known as hard attention or non-deterministic attention, is an alignment type that forces the attention model to focus on exactly one feature vector. Firstly, this method implements the softmax function in the exact same way as global alignment. However, the outputs $a_{1}, \dots, a_{n_f}$ are not used as weights for the context vector calculation. Instead, these values are used as probabilities to draw the choice of the one value vector from. A value $m \in \mathbb{R}^{1}$ is drawn from a multinomial distribution with $a_{1}, \dots, a_{n_f}$ as parameters for the probabilities. Then, the context vector is simply defined as follows:
\begin{equation}
    \def\sss{\scriptscriptstyle}
    \setstackgap{L}{8pt}
    \def\stacktype{L}
    \stackunder{\bm c}{\sss d_v \times 1} = \stackunder{\bm v_{m}}{\sss d_v \times 1}.
\end{equation}

Hard alignment is typically more efficient at inference compared to soft alignment. On the other hand, the main disadvantage of hard attention is that, due to the stochastic alignment of attention, the training of the model cannot be done via the regular backpropagation method. Instead, simulation and sampling, or reinforcement learning \cite{Williams1992} are required to calculate the gradient at the hard attention layer. As such, soft/global attention is generally preferred. However, a compromise can be made in certain situations. \textbf{Local alignment} \cite{luong2015effective} is a method that implements a softmax distribution, similarly to soft/global alignment. But, the softmax distribution is calculated based only on a subset of the inputs. This method is generally used in combination with sequence data. One has to specify a variable $p \in \mathbb{R}^{1}$ that determines the position of the region. Feature vectors close to $p$ will be attended to by the model, and vectors too far from $p$ will be ignored. The size of the subset will be determined by the variable $D \in \mathbb{R}^{1}$. Summarizing, the attention model will apply a softmax function on the attention scores in the subset $[p - D, p + D]$. In other words, a window is placed on the input and soft/global attention is calculated within that window:
\begin{equation}\label{equation:LocalAttentionWeight}
    \def\sss{\scriptscriptstyle}
    \setstackgap{L}{8pt}
    \def\stacktype{L}
    \stackunder{a_{l}}{\sss 1 \times 1} = \text{align}(\stackunder{e_{l}}{\sss 1 \times 1}; \stackunder{\bm e}{\sss n_f \times 1}) = \frac{\text{exp}(e_{l})}{\sum^{p+D}_{j=p-D}\text{exp}(e_{j})}.
\end{equation}
The question that remains is how to determine the location parameter $p$. The first method is referred to as \textbf{monotonic alignment}. This straightforward method entails simply setting the location parameter equal to the location of the prediction in the output sequence. Another method of determining the position of the region is referred to as \textbf{predictive alignment}. As the name entails, the model attempts to actually predict the location of interest in the sequence:
\begin{equation}\label{equation:PredictiveAlignment}
    \def\sss{\scriptscriptstyle}
    \setstackgap{L}{8pt}
    \def\stacktype{L}
    \stackunder{p}{\sss 1 \times 1} = \stackunder{S}{\sss 1 \times 1} \times \text{sigmoid}(\stackunder{\bm w_p^T}{\sss 1 \times d_p} \times \text{tanh}(\stackunder{\bm W_p}{\sss d_p \times d_q} \times  \stackunder{\bm q}{\sss d_q \times 1})),
\end{equation}
where $S \in \mathbb{R}^{1}$ is the length of the input sequence, and $\bm w_p \in \mathbb{R}^{d_{p}}$ and $\bm W_p \in \mathbb{R}^{d_{p} \times d_q}$ are both trainable weights parameters. The sigmoid function multiplied by $S$ makes sure that $p$ is in the range $[0, S]$. Additionally, in \cite{luong2015effective}, it is recommended to add an additional term to the alignment function to favor alignment around $p$:
\begin{equation}\label{equation:LocalAttentionWeightAdjusted}
    \def\sss{\scriptscriptstyle}
    \setstackgap{L}{8pt}
    \def\stacktype{L}
    \stackunder{a_{l}}{\sss 1 \times 1} = \text{align}(\stackunder{e_{l}}{\sss 1 \times 1}; \stackunder{\bm e}{\sss n_f \times 1})\text{exp}(-\frac{(l-p)^2)}{2 \sigma ^2}),
\end{equation}
where $\sigma \in \mathbb{R}^{1}$ is empirically set equal to $\frac{D}{2}$ according to \cite{luong2015effective}. Another proposed method for compromising between soft and hard alignment is \textbf{reinforced alignment} \cite{DBLP:conf/ijcai/ShenZLJWZ18}. Similarly to local alignment, a subset of the feature vectors is determined, for which soft alignment is calculated. However, instead of using a window to determine the subset, reinforced alignment uses a reinforcement learning agent \cite{Williams1992}, similarly to hard alignment, to choose the subset of feature vectors. The attention calculation based on these chosen feature vectors is the same as regular soft alignment.

Soft alignment is often regarded as the standard alignment function for attention models in practically every domain. Yet, the other alignment methods have also seen interesting uses in various domains. For example, hard attention is used in \cite{Malinowski_2018_ECCV} for the task of visual question answering. In \cite{liu2020multi}, both soft and hard attention are used in a graph attention model for multi-agent game abstraction. Similarly, in \cite{10.1145/3109859.3109890}, both global and local alignment are used for review rating predictions. Reinforced alignment has been employed in combination with a co-attention structure in \cite{wang-etal-2019-aspect} for the task of aspect sentiment classification. In \cite{JIANG2020102775}, reinforced alignment is used for the task of person re-identification using surveillance images.

\subsubsection{Attention Dimensionality}\label{sec:Dimensionality}
All previous model specifications of attention use a scalar weight $a_{l}$ for each value vector $\bm v_l$. This technique is referred to as \textbf{single-dimensional attention}. However, instead of determining a single attention score and weight for the entire vector, \cite{shen2018disan} proposes to calculate weights for every single feature in those vectors separately. This technique is referred to as \textbf{multi-dimensional attention}, since the attention weights now become higher dimensional vectors. The idea is that the model no longer has to attend to entire vectors, but it can instead pick and choose specific elements from those vectors. More specifically, attention is calculated for each dimension. As such, the model must create a vector of attention weights $\bm a_{l} \in \mathbb{R}^{d_v}$ for each value vector $\bm v_l \in \mathbb{R}^{d_v}$. The context vector can then be calculated by summing the element-wise multiplications ($ \circ $) of the value vectors $\bm v_1, \dots, \bm v_{n_f} \in \mathbb{R}^{d_v}$ and the corresponding attention weight vectors $\bm a_{1}, \dots, \bm a_{n_f} \in \mathbb{R}^{d_v}$, as follows:
\begin{equation}\label{equation:MultidimensionalContextVector}
    \def\sss{\scriptscriptstyle}
    \setstackgap{L}{8pt}
    \def\stacktype{L}
    \stackunder{\bm c}{\sss d_v \times 1} = \sum^{n_f}_{l=1} \stackunder{\bm a_{l}}{\sss d_v \times 1} \circ \stackunder{\bm v_l}{\sss d_v \times 1}.
\end{equation}
However, since one needs to create attention weight vectors, this technique requires adjusted attention score and weight calculations. For example, the concatenate score function found in Table \ref{Table:ScoreFunctions} can be adjusted by changing the $\bm w \in \mathbb{R}^{d_w}$ weights vector to the weight matrix $\bm W_d \in \mathbb{R}^{d_w \times d_v}$:
\begin{equation}
\def\sss{\scriptscriptstyle}
    \setstackgap{L}{8pt}
    \def\stacktype{L}
    \stackunder{\bm e_{l}}{\sss d_v \times 1} = \stackunder{\bm W_d^T}{\sss d_v \times d_w} \times  \text{act}(\stackunder{\bm W_1}{\sss d_w \times d_q} \times \stackunder{\bm q}{\sss d_q \times 1} + \stackunder{\bm W_2}{\sss d_w \times d_k} \times \stackunder{\bm k_l}{\sss d_k \times 1} + \stackunder{\bm b}{\sss d_w \times 1} ).
\end{equation}
This new score function produces the attention score vectors $\bm e_{1}, \dots, \bm e_{n_f} \in \mathbb{R}^{d_v}$. These score vectors can be combined into a matrix of scores $\bm e = [\bm e_{1}, \dots, \bm e_{n_f}] \in \mathbb{R}^{d_v \times n_f}$. To produce multi-dimensional attention weights, the alignment function stays the same, but it is applied for each feature across the attention score columns. To illustrate, when implementing soft attention, the attention weight produced from the $i$th element of score vector $\bm e_{l}$ is defined as follows:
\begin{equation}\label{equation:MultidimensionalAttentionWeights}
    \def\sss{\scriptscriptstyle}
    \setstackgap{L}{8pt}
    \def\stacktype{L}
    \stackunder{a_{l,i}}{\sss 1 \times 1} = \text{align}(\stackunder{e_{l,i}}{\sss 1 \times 1}; \stackunder{\bm e}{\sss d_v \times n_f}) = \frac{\text{exp}(e_{l,i})}{\sum^{n_f}_{j=1}\text{exp}(e_{j,i})},
\end{equation}
where $e_{l,i}$ represents the $i$th element of score vector $\bm e_{l}$, and $a_{l,i}$ is the $i$th element of the attention weights vector $\bm a_{l}$. Finally, these attention weight vectors can be used to compute the context vector as presented in (\ref{equation:MultidimensionalContextVector}).

Multi-dimensional attention is a very general mechanism that can be applied in practically every attention model, but actual applications of the technique have been relatively sparse. One application example is \cite{arshad2019aiding}, where multi-dimensional attention is used in a model for named entity recognition based on text and visual context from multimedia posts. In \cite{wu-etal-2018-question}, multi-dimensional attention is used in a model for answer selection in community question answering. In \cite{oktay2018attention}, the U-net model for medical image segmentation is extended with a multi-dimensional attention mechanism. Similarly, in \cite{10.1007/978-3-030-32236-6_16}, the Transformer model is extended with the multi-dimensional attention mechanism for the task of dialogue response generation. In \cite{chen2020schema}, multi-dimensional attention is used to extend graph attention networks for dialogue state tracking. Lastly, for the task of next-item recommendation, \cite{ijcai2019-513} proposes a model that incorporates multi-dimensional attention.

\subsection{Query-Related Attention Mechanisms}\label{sec:ModelOutput}
Queries are an important part of any attention model, since they directly determine which information is extracted from the feature vectors. These queries are based on the desired output of the task model, and can be interpreted as literal questions. Some queries have specific characteristics that require specific types of mechanisms to process them. As such, this category encapsulates the attention mechanisms that deal with specific types of query characteristics. The mechanisms in this category deal with one of the two following query characteristics: the type of queries or the multiplicity of queries.

\subsubsection{Type of Queries}\label{sec:OutputTypes}
Different attention models employ attention for different purposes, meaning that distinct query types are necessary. There are \textbf{basic queries}, which are queries that are typically straightforward to define based on the data and model. For example, the hidden state for one prediction in an RNN is often used as the query for the next prediction. One could also use a vector of auxiliary variables as query. For example, when doing medical image classification, general patient characteristics can be incorporated into a query.

Some attention mechanisms, such as co-attention, rotatory attention, and attention-over-attention, use \textbf{specialized queries}. For example, rotatory attention uses the context vector from another attention module as query, while interactive co-attention uses an averaged keys vector based on another input. Another case one can consider is when attention is calculated based purely on the feature vectors. This concept has been mentioned before and is referred to as \textbf{self-attention} or \textbf{intra-attention} \cite{lin2017structured}. We say that the models use \textbf{self-attentive queries}. There are two ways of interpreting such queries. Firstly, one can say that the query is constant. For example, document classification requires only a single classification as the output of the model. As such, the query is always the same, namely: ``What is the class of the document?". The query can be ignored and attention can be calculated based only on the features themselves. Score functions can be adjusted for this by making the query vector a vector of constants or removing it entirely:
\begin{equation}\label{equation:Self-AttentiveAdd}
\def\sss{\scriptscriptstyle}
\setstackgap{L}{11pt}
\def\stacktype{L}
    \text{score}(\hspace{-3pt}\stackunder{\bm k_l}{\sss d_k \times 1}\hspace{-3pt}) = \stackunder{\bm w^T}{\sss 1 \times d_w} \hspace{-4pt}\times \text{act}(\hspace{-4pt}\stackunder{\bm W}{\sss d_w \times d_{k}} \hspace{-4pt}\times \stackunder{\bm k_{l}}{\sss d_{k} \times 1}\hspace{-3pt} + \hspace{-6pt}\stackunder{\bm b}{\sss d_{w} \times 1}\hspace{-4pt}).
\end{equation}
Additionally, one can also interpret self-attention as learning the query along the way, meaning that the query can be defined as a trainable vector of weights. For example, the dot-product score function may take the following form:
\begin{equation}\label{equation:Self-AttentiveDot}
\def\sss{\scriptscriptstyle}
\setstackgap{L}{11pt}
\def\stacktype{L}
    \text{score}(\hspace{-3pt}\stackunder{\bm k_l}{\sss d_k \times 1}\hspace{-3pt}) = \stackunder{\bm q^T}{\sss 1 \times d_k} \hspace{-6pt}\times\hspace{-3pt} \stackunder{\bm k_{l}}{\sss d_{k} \times 1},
\end{equation}
where $\bm q \in \mathbb{R}^{d_k}$ is a trainable vector of weights. One could also interpret vector $\bm b \in \mathbb{R}^{d_w}$ as the query in (\ref{equation:Self-AttentiveAdd}). Another use of self-attention is to uncover the relations between the feature vectors $\bm f_1, \dots, \bm f_{n_f}$. These relations can then be used as additional information to incorporate into new representations of the feature vectors. With basic attention mechanisms, the keys matrix $\bm K$, and the values matrix $\bm V$ are extracted from the features matrix $\bm F$, while the query $\bm q$ is produced separately. For this type of self-attention, the query vectors are extracted in a similar process as the keys and values, via a transformation matrix of trainable weights $\bm W_Q \in \mathbb{R}^{d_q \times d_f}$. We define the matrix $\bm Q = [\bm q_1, \dots, \bm q_{n_f}] \in \mathbb{R}^{d_q \times n_f}$, which can be obtained as follows:
\begin{equation}\label{equation:SelfAttentionQueries}
    \def\sss{\scriptscriptstyle}
    \setstackgap{L}{8pt}
    \def\stacktype{L}
    \stackunder{\bm Q}{\sss d_q \times n_f} = \stackunder{\bm W_Q}{\sss d_q \times d_f} \times \stackunder{\bm F}{\sss d_f \times n_f}.
\end{equation}

Each column of $\bm Q$ can be used as the query for the attention model. When attention is calculated using a query $\bm q$, the resulting context vector $\bm c$ will summarize the information in the feature vectors that is important to the query. Since the query, or a column of $\bm Q$, is now also a feature vector representation, the context vector contains the information of all feature vectors that are important to that specific feature vector. In other words, the context vectors capture the relations between the feature vectors. For example, self-attention allows one to extract the relations between words: which verbs refer to which nouns, which pronouns refer to which nouns, etc. For images, self-attention can be used to determine which image regions relate to each other.

While self-attention is placed in the query-related category, it is also very much related to the feature model. Namely, self-attention is a technique that is often used in the feature model to create improved representations of the feature vectors. For example, the Transformer model for language processing \cite{vaswani2017attention}, and the Transformer model for image processing \cite{parmar2018image}, both use multiple rounds of (multi-head) self-attention to improve the representation of the feature vectors. The relations captured by the self-attention mechanism are incorporated into new representations. A simple method of determining such a new representation is to simply set the feature vectors equal to the acquired self-attention context vectors \cite{lin2017structured}, as presented in (\ref{equation:SelfAttentionRepresentation1}).
\begin{equation}\label{equation:SelfAttentionRepresentation1}
    \def\sss{\scriptscriptstyle}
    \setstackgap{L}{8pt}
    \def\stacktype{L}
    \stackunder{\bm f^{(\text{new})}}{\sss d_f \times 1} = \stackunder{\bm c}{\sss d_f \times 1},
\end{equation}
where $\bm f^{(\text{new})}$ is the updated feature vector. Another possibility is to add the context vectors to the previous feature vectors with an additional normalization layer \cite{vaswani2017attention}:
\begin{equation}\label{equation:SelfAttentionRepresentation2}
    \def\sss{\scriptscriptstyle}
    \setstackgap{L}{8pt}
    \def\stacktype{L}
    \stackunder{\bm f^{(\text{new})}}{\sss d_f \times 1} = \text{Normalize}(\stackunder{\bm f^{(\text{old})}}{\sss d_f \times 1} + \stackunder{\bm c}{\sss d_f \times 1}),
\end{equation}
where $\bm f^{(\text{old})}$ is the previous feature vector, and $\text{Normalize}()$ is a normalization layer \cite{ba2016layer}. Using such techniques, self-attention has been used to create improved word or sentence embeddings that enhance model accuracy \cite{lin2017structured}.

Self-attention is arguably one of the more important types of attention, partly due to its vital role in the highly popular Transformer model. Self-attention is a very general mechanism and can be applied to practically any problem. As such, self-attention has been extensively explored in many different fields in both Transformer-based architectures and other types of models. For example, in \cite{Zhao_2020_CVPR}, self-attention is explored for image recognition tasks, and results indicate that the technique may have substantial advantages with regards to robustness and generalization. In \cite{zhang2018selfattention}, self-attention is used in a generative adversarial network (GAN) \cite{goodfellow2014generative} to determine which regions of the input image to focus on when generating the regions of a new image. In \cite{9066969}, self-attention is used to design a state-of-the-art medical image segmentation model. Naturally, self-attention can also be used for video processing. In \cite{fajtl2018summarizing}, a self-attention model is proposed for the purpose of video summarization that reaches state-of-the-art results. In other fields, like audio processing, self-attention has been explored as well. In \cite{salazar2019self}, self-attention is used to create a speech recognition model. Self-attention has also been explored in overlapping domains. For example, in \cite{afouras2018deep}, the self-attention Transformer architecture is used to create a model that can recognize phrases from audio and by lip-reading from a video. For the problem of next item recommendation, \cite{zhang2018next} proposes a Transformer model that explicitly captures item-item relations using self-attention. Self-attention also has applications in any natural language processing fields. For example, in \cite{letarte2018importance}, self-attention is used for sentiment analysis. Self-attention is also highly popular for graph models. For example, self-attention is explored in \cite{sankar2020dysat} for the purpose of representation learning in communication networks and rating networks. Additionally, the first attention model for graph networks was based on self-attention \cite{velikovi2017graph}.

\subsubsection{Multiplicity of Queries}\label{sec:OutputCount}
In previous examples, the attention model generally used a single query for a prediction. We say that such models use \textbf{singular query attention}. However, there are attention architectures that allow the model to compute attention using multiple queries. Note that this is different from, for example, an RNN that may involve multiple queries to produce a sequence of predictions. Namely, such a model still requires only a single query per prediction.

\begin{figure}
    \centering
    \includegraphics[scale=0.6]{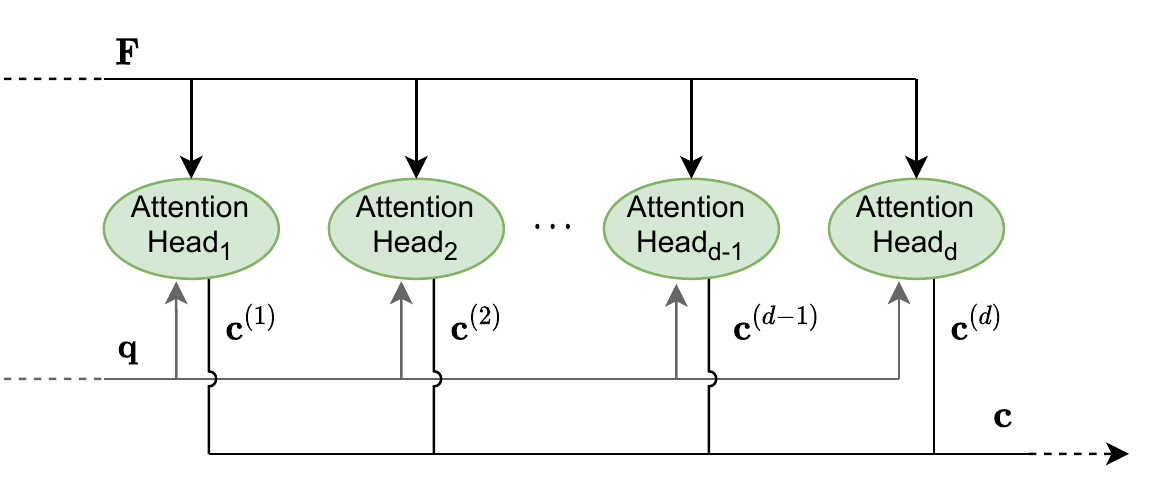}
    \caption{An illustration of multi-head attention.}
    \label{fig:MultiheadAttentionModule}
\end{figure}

One example of a technique that incorporates multiple queries is \textbf{multi-head attention} \cite{vaswani2017attention}, as presented in Fig. \ref{fig:MultiheadAttentionModule}. Multi-head attention works by implementing multiple attention modules in parallel by utilizing multiple different versions of the same query. The idea is to linearly transform the query $\bm q$ using different weight matrices. Each newly formed query essentially asks for a different type of relevant information, allowing the attention model to introduce more information into the context vector calculation. An attention model implements $d \geq 1$ heads with each attention head having its own query vector, keys matrix, and values matrix: $\bm q^{(j)}$, $\bm K^{(j)}$ and $\bm V^{(j)}$, for $j=1, \dots, d$. The query $\bm q^{(j)}$ is obtained by linearly transforming the original query $\bm q$, while the matrices $\bm K^{(j)}$ and $\bm V^{(j)}$ are obtained through linear transformations of $\bm F$. As such, each attention head has its own learnable weights matrices $\bm W_q^{(j)}$, $\bm W_K^{(j)}$ and $\bm W_V^{(j)}$ for these transformations. The calculation of the query, keys, and values for the $j$th head are defined as follows:
\begin{equation}\label{equation:MultiHeadQueries}
    \def\sss{\scriptscriptstyle}
    \setstackgap{L}{8pt}
    \def\stacktype{L}
\begin{split}
    \stackunder{\bm q^{(j)}}{\sss d_q \times 1} = \stackunder{\bm W_q^{(j)}}{\sss d_q \times d_q} \times \stackunder{\bm q}{\sss d_q \times 1},
    \hspace{30pt}
    \stackunder{\bm K^{(j)}}{\sss d_k \times n_f} = \stackunder{\bm W_K^{(j)}}{\sss d_k \times d_f} \times \stackunder{\bm F}{\sss d_f \times n_f}, \\
    \stackunder{\bm V^{(j)}}{\sss d_v \times n_f} = \stackunder{\bm W_V^{(j)}}{\sss d_v \times d_f} \times \stackunder{\bm F}{\sss d_f \times n_f}. \hspace{60pt}
\end{split}
\end{equation}
Thus, each head creates its own representations of the query $\bm q$, and the input matrix $\bm F$. Each head can therefore learn to focus on different parts of the inputs, allowing the model to attend to more information. For example, when training a machine translation model, one attention head can learn to focus on which nouns (e.g., student, car, apple)  do certain verbs (e.g., walking, driving, buying) refer to, while another attention head learns to focus on which nouns refer to certain pronouns (e.g., he, she, it) \cite{vaswani2017attention}. Each head will also create its own vector of attention scores $\bm e^{(j)} = [e_{1}^{(j)}, \dots, e_{n_f}^{(j)}]\in \mathbb{R}^{n_f}$, and a corresponding vector of attention weights $\bm a^{(j)} = [a_{1}^{(j)}, \dots, a_{n_f}^{(j)}]\in \mathbb{R}^{n_f}$. As can be expected, each attention model produces its own context vector $\bm c^{(j)} \in \mathbb{R}^{d_v}$, as follows:
\begin{equation}\label{equation:ContextHead}
    \def\sss{\scriptscriptstyle}
    \setstackgap{L}{8pt}
    \def\stacktype{L}
    \stackunder{\bm c^{(j)}}{\sss d_v \times 1} = \sum^{n_f}_{l=1} \stackunder{a_{l}^{(j)}}{\sss 1 \times 1} \times \stackunder{\bm v_l^{(j)}}{\sss d_v \times 1}.
\end{equation}

The goal is still to create a single context vector as output of the attention model. As such, the context vectors produced by the individual attention heads are concatenated into a single vector. Afterwards, a linear transformation is applied using the weight matrix $\bm W_O \in \mathbb{R}^{d_c \times d_v d}$ to make sure the resulting context vector $\bm c \in \mathbb{R}^{d_c}$ has the desired dimension. This calculation is presented in (\ref{equation:ContextHead}). The dimension $d_c$ can be pre-specified by, for example, setting it equal to $d_v$, so that the context vector dimension is unchanged.
\begin{equation}\label{equation:ContextHead}
    \def\sss{\scriptscriptstyle}
    \setstackgap{L}{8pt}
    \def\stacktype{L}
    \stackunder{\bm c}{\sss d_c \times 1} = \stackunder{\bm W_O}{\sss d_c \times d_v d} \times \text{concat}(\stackunder{\bm c^{(1)}}{\sss d_v \times 1},...,\stackunder{\bm c^{(d)}}{\sss d_v \times 1}).
\end{equation}

Multi-head attention processes multiple attention modules in parallel, but attention modules can also be implemented sequentially to iteratively adjust the context vectors. Each of these attention modules are referred to as ``repetitions" or ``rounds" of attention. Such attention architectures are referred to as \textbf{multi-hop attention models}, also known as \textbf{multi-step attention models}. An important note to consider is the fact that multi-hop attention is a mechanism that has been proposed in various forms throughout various works. While the mechanism always involves multiple rounds of attention, the multi-hop implementation proposed in \cite{iida2019attention} differs from the mechanism proposed in \cite{tran2018multihop} or \cite{gong2017ruminating}. Another interesting example is \cite{yoon2019speech}, where a ``multi-hop" attention model is proposed that would actually be considered alternating co-attention in this survey, as explained in Subsection \ref{sec:InputCount}.

\begin{figure}
    \centering
    \includegraphics[scale=0.6]{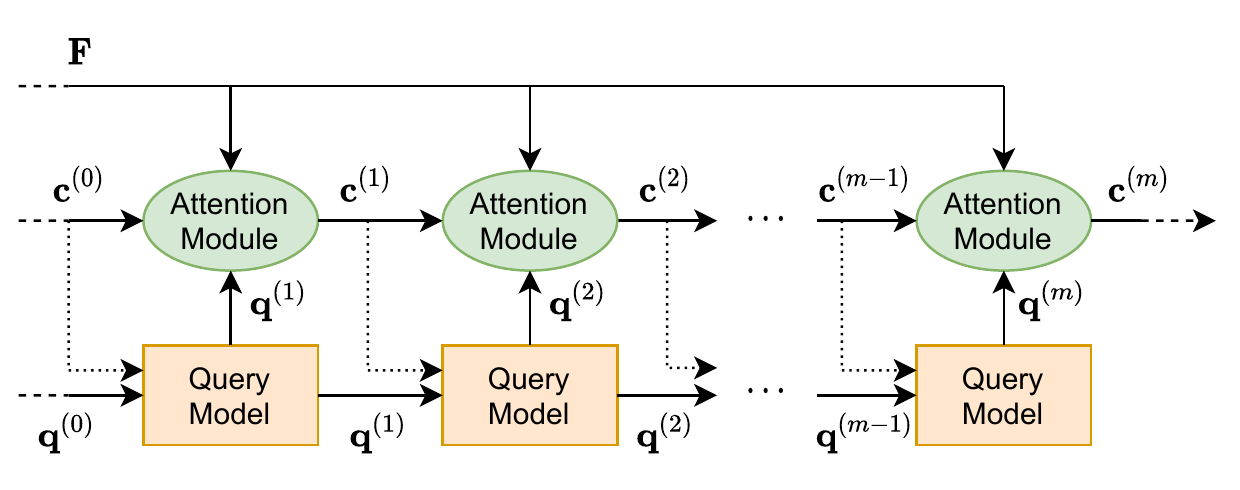}
    \caption{An example illustration of multi-hop attention. Solid arrows represent the base multi-hop model structure, while dotted arrows represent optional connections.}
    \label{fig:MultihopAttentionModule}
\end{figure}

We present a general form of multi-hop attention that is largely a generalization of the techniques introduced in \cite{tran2018multihop} and \cite{yang2016stacked}. Fig. \ref{fig:MultihopAttentionModule} provides an example implementation of a multi-hop attention mechanism.
The general idea is to iteratively transform the query, and use the query to transform the context vector, such that the model can extract different information in each step. Remember that a query is similar to a literal question. As such, one can interpret the transformed queries as asking the same question in a different manner or from a different perspective, similarly to the queries in multi-head attention. The query that was previously denoted by $\bm q$ is now referred to as the initial query, and is denoted by $\bm q^{(0)}$. At hop $s$, the current query $\bm q^{(s)}$ is transformed into a new query representation $\bm q^{(s+1)}$, possibly using the current context vector $\bm c^{(s)}$ as another input, and some transformation function $\text{transform}()$:
\begin{equation}
\def\sss{\scriptscriptstyle}
\setstackgap{L}{8pt}
\def\stacktype{L}
\stackunder{\bm q^{(s+1)}}{\sss d_q \times 1} = \text{transform}(\stackunder{\bm q^{(s)}}{\sss d_q \times 1},\; \stackunder{\bm c^{(s)}}{\sss d_v \times 1}).
\end{equation}
For the specific form of the transformation function $\text{transform}()$, \cite{tran2018multihop} proposes to use a mechanism similar to self-attention. 
Essentially, the queries used by the question answer matching model proposed in \cite{tran2018multihop} were originally based on a set of feature vectors extracted from a question. \cite{tran2018multihop} also defines the original query $\bm q^{(0)}$ as the unweighted average of these feature vectors. At each hop $s$, attention can be calculated on these feature vectors using the previous query $\bm q^{(s)}$ as the query in this process. The resulting context vector of this calculation is the next query vector. Using the context vector $\bm c^{(s)}$ instead of $\bm q^{(s)}$ as the query for this process is also a possibility, which is similar to the \textit{LCR-Rot-hop} model proposed in \cite{wallaart2019hybrid} and the multi-step model proposed in \cite{yang2016stacked}. Such a connection is represented by the dotted arrows in Fig. \ref{fig:MultihopAttentionModule}. The transformation mechanism uses either the $\bm q^{(s)}$ or the context vector $\bm c^{(s)}$ as query, but a combination via concatenation is also possible.

Each query representation is used as input for the attention module to compute attention on the columns of the feature matrix $\bm F$, as seen previously. One main difference, however, is that the context vector $\bm c^{(s)}$ is also used as input, so that the actual query input for the attention model is the concatenation of $\bm c^{(s)}$ and $\bm q^{(s+1)}$. The adjusted attention score function is presented in (\ref{equation:MultihopAttentionScore}). Note that the initial context vector $\bm c^{(0)}$ is predefined. One way of doing this is by setting it equal to the unweighted average of the value vectors $\bm v_1, \dots, \bm v_{n_f} \in \mathbb{R}^{d_v}$ extracted from $\bm F$.
\begin{equation}\label{equation:MultihopAttentionScore}
    \def\sss{\scriptscriptstyle}
    \setstackgap{L}{8pt}
    \def\stacktype{L}
   \stackunder{e_{l}^{(s)}}{\sss 1 \times 1} = \text{score}(\text{concat}(\stackunder{\bm q^{(s+1)}}{\sss d_q \times 1}, \stackunder{\bm c^{(s)}}{\sss d_v \times 1} ), \stackunder{\bm k_l}{\sss d_k \times 1}\hspace{-2pt}).
\end{equation}
An alignment function and the value vectors are then used to produce the next context vector $\bm c^{(s+1)}$. One must note that in \cite{tran2018multihop}, the weights used in each iteration are the same weights, meaning that the number of parameters do not scale with the number of repetitions. Yet, using multiple hops with different weight matrices can also be viable, as shown by the Transformer model \cite{vaswani2017attention} and in \cite{yang2016stacked}. It may be difficult to grasp why $\bm c^{(s)}$ is part of the query input for the attention model. Essentially, this technique is closely related to self-attention in the sense that, in each iteration, a new context representation is created from the feature vectors and the context vector. The essence of this mechanism is that one wants to iteratively alter the query and the context vector, while attending to the feature vectors. In the process, the new representations of the context vector absorb more different kinds of information. This is also the main difference between this type of attention and multi-head attention. Multi-head attention creates multiple context vectors from multiple queries and combines them to create a final context vector as output. Multi-hop attention iteratively refines the context vector by incorporating information from the different queries. This does have the disadvantage of having to calculate attention sequentially.

Interestingly, due to the variations in which multi-hop attention has been proposed, some consider the Transformer model's encoder and decoder to consist of several single-hop attention mechanisms \cite{iida2019attention} instead of being a multi-hop model. However, in the context of this survey, we consider the Transformer model to be an alternative form of the multi-hop mechanism, as the features matrix $\bm F$ is not directly reused in each step. Instead, $\bm F$ is only used as an input for the first hop, and is transformed via self-attention into a new representation. The self-attention mechanism uses each feature vector in $\bm F$ as a query, resulting in a matrix of context vectors as output of each attention hop. The intermediate context vectors are turned into matrices and represent iterative transformations of the matrix $\bm F$, which are used in the consecutive steps. Thus, the Transformer model iteratively refines the features matrix $\bm F$ by extracting and incorporating new information.

\begin{figure}
    \centering
    \includegraphics[scale=0.6]{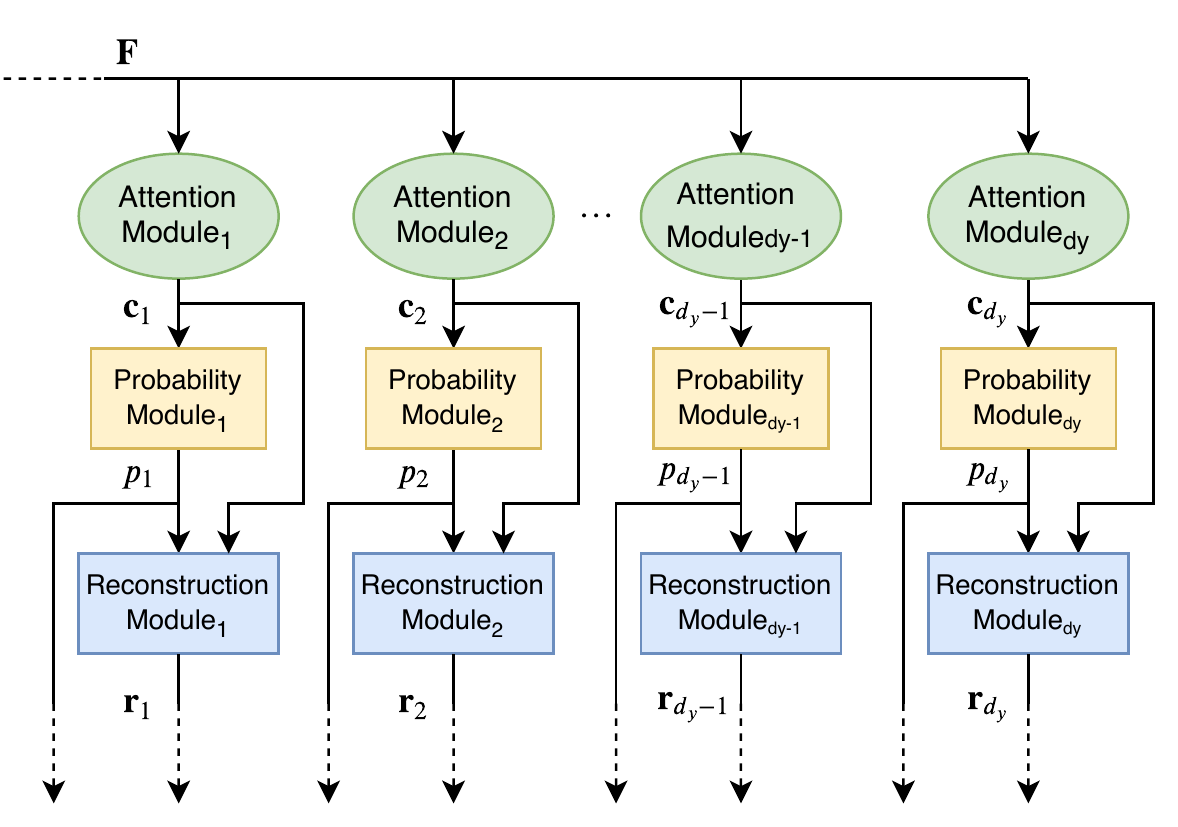}
    \caption{An illustration of capsule-based attention.}
    \label{fig:CapsuleAttention}
\end{figure}

When dealing with a classification task, another idea is to use a different query for each class. This is the basic principle behind \textbf{capsule-based attention} \cite{10.1145/3178876.3186015}, as inspired by the capsule networks \cite{sabour2017dynamic}. Suppose we have the feature vectors $\bm f_1, \dots , \bm f_{n_f} \in \mathbb{R}^{d_{f}}$, and suppose there are are $d_y$ classes that the model can predict. Then, a capsule-based attention model defines a capsule for each of the $d_y$ classes that each take as input the feature vectors. Each capsule consists of, in order, an attention module, a probability module, and a reconstruction module, which are depicted in Fig. \ref{fig:CapsuleAttention}. The attention modules all use self-attentive queries, so each module learns its own query: "Which feature vectors are important to identify this class?". In \cite{10.1145/3178876.3186015}, a self-attentive multiplicative score function is used for this purpose:
\begin{equation}\label{equation:CapsuleDot}
\def\sss{\scriptscriptstyle}
\setstackgap{L}{11pt}
\def\stacktype{L}
    \stackunder{e_{c,l}}{\sss 1 \times 1} = \stackunder{\bm q_c^T}{\sss 1 \times d_k} \times \stackunder{\bm k_{l}}{\sss d_{k} \times 1},
\end{equation}
where $e_{c,l} \in \mathbb{R}^{1}$ is the attention score for vector $l$ in capsule $c$, and $\bm q_c \in \mathbb{R}^{d_{k}}$ is a trainable query for capsule $c$, for $c = 1, \dots, d_y$. Each attention module then uses an alignment function, and uses the produced attention weights to determine a context vector $\bm c_c \in \mathbb{R}^{d_{v}}$. Next, the context vector $\bm c_c$ is fed through a probability layer consisting of a linear transformation with a sigmoid activation function:
\begin{equation}\label{equation:CapsuleProb}
\def\sss{\scriptscriptstyle}
\setstackgap{L}{11pt}
\def\stacktype{L}
    \stackunder{p_{c}}{\sss 1 \times 1} = \text{sigmoid}(\stackunder{\bm w_c^T}{\sss 1 \times d_v} \hspace{-3pt}\times\hspace{-3pt} \stackunder{\bm c_{c}}{\sss d_{v} \times 1} + \stackunder{b_c}{\sss 1 \times 1}),
\end{equation}
where $\bm w_c \in \mathbb{R}^{d_{v}}$ and $b_c \in \mathbb{R}^{1}$ are trainable capsule-specific weights parameters, and $p_c \in \mathbb{R}^{1}$ is the predicted probability that the correct class is class $c$. The final layer is the reconstruction module that creates a class vector representation. This representation $\bm r_c \in \mathbb{R}^{d_v}$ is determined by simply multiplying the context vector $\bm c_c$ by the probability $p_c$:
\begin{equation}\label{equation:CapsuleRepres}
\def\sss{\scriptscriptstyle}
\setstackgap{L}{11pt}
\def\stacktype{L}
    \stackunder{\bm r_{c}}{\sss d_v \times 1} = \stackunder{p_c}{\sss 1 \times 1} \hspace{-1pt}\times\hspace{-1pt} \stackunder{\bm c_{c}}{\sss d_{v} \times 1}.
\end{equation}
The capsule representation is used when training the model. First of all, the model is trained to predict the probabilities $p_1, \dots, p_{d_y}$ as accurately as possible compared to the true values. Secondly, via a joint loss function, the model is also trained to accurately construct the capsule representations $\bm r_{1}, \dots, \bm r_{d_y}$. A features representation $\bm f \in \mathbb{R}^{d_f}$ is defined which is simply the unweighted average of the original feature vectors. The idea is to train the model such that vector representations from capsules that are not the correct class differ significantly from $\bm f$ while the representation from the correct capsule is very similar to $\bm f$. A dot-product between the capsule representations and the features representation is used in \cite{10.1145/3178876.3186015} as a measure of the distance between the vectors. Note that $d_v$ must equal $d_f$ in this case, otherwise the vectors would have incompatible dimensions. Interestingly, since attention is calculated for each class individually, one can track which specific feature vectors are important for which specific class. In \cite{10.1145/3178876.3186015}, this idea is used to discover which words correspond to which sentiment class.

The number of tasks that can make use of multiple queries is substantial, due to how general the mechanisms are. As such, the techniques described in this section have been extensively explored in various domains. For example, multi-head attention has been used for speaker recognition based on audio spectrograms \cite{india2019self}. In \cite{wu-etal-2019-neural-news}, multi-head attention is used for recommendation of news articles. Additionally, multi-head attention can be beneficial for graph attention models as well \cite{velikovi2017graph}. As for multi-hop attention, quite a few papers have been mentioned before, but there are still many other interesting examples. For example, in \cite{wang2021adversarially}, a multi-hop attention model is proposed for medication recommendation. Furthermore, practically every Transformer model makes use of both multi-head and multi-hop attention. The Transformer model has been extensively explored in various domains. For example, in \cite{cornia2020meshed}, a Transformer model is implemented for image captioning. In \cite{chen2021transunet}, Transformers are explored for medical image segmentation. In \cite{zhong-etal-2019-knowledge}, a Transformer model is used for emotion recognition in text messages. A last example of an application of Transformers is \cite{sun2019bert4rec}, which proposes a Transformer model for recommender systems. In comparison with multi-head and multi-hop attention, capsule-based attention is arguably the least popular of the mechanisms discussed for the multiplicity of queries. One example is \cite{zhou2019dynamic}, where an attention-based capsule network is proposed that also includes a multi-hop attention mechanism for the purpose of visual question answering. Another example is \cite{wang2019aspect}, where capsule-based attention is used for aspect-level sentiment analysis of restaurant reviews.

The multiplicity of queries is a particularly interesting category due to the Transformer model \cite{vaswani2017attention}, which combines a form of multi-hop and multi-head attention. Due to the initial success of the Transformer model, many improvements and iterations of the model have been produced that typically aim to improve the predictive performance, the computational efficiency, or both. For example, the Transformer-XL \cite{DBLP:conf/acl/DaiYYCLS19} is an extension of the original Transformer that uses a recurrence mechanism to not be limited by a context window when processing the outputs. This allows the model to learn significantly longer dependencies while also being computationally more efficient during the evaluation phase. Another extension of the Transformer is known as the Reformer model \cite{kitaev2020reformer}. This model is significantly more efficient computationally, by means of locality-sensitive hashing, and memory-wise, by means of reversible residual layers. Such computational improvements are vital, since one of the main disadvantages of the Transformer model is the sheer computational cost due to the complexity of the model scaling quadratically with the amount of input feature vectors. The Linformer model \cite{wang2020linformer} manages to reduce the complexity of the model to scale linearly, while achieving similar performance as the Transformer model. This is achieved by approximating the attention weights using a low-rank matrix. The Lite-Transformer model proposed in \cite{wu2020lite} achieves similar results by implementing two branches within the Transformer block that specialize in capturing global and local information. Another interesting Transformer architecture is the Synthesizer \cite{tay2020synthesizer}. This model replaces the pairwise self-attention mechanism with ``synthetic" attention weights. Interestingly, the performance of this model is relatively close to the original Transformer, meaning that the necessity of the pairwise self-attention mechanism of the Transformer model may be questionable. For a more comprehensive overview of Transformer architectures, we refer to \cite{tay2020efficient}.

\begin{table*}
\caption{Attention models analyzed based on the proposed taxonomy. A plus sign (+) between two mechanisms indicates that both techniques were combined in the same model, while a comma (,) indicates that both mechanisms were tested in the same paper, but not necessarily as a combination in the same model.}
\begin{center}
\resizebox{\linewidth}{!}{%
\begin{tabular}{|p{100pt}||p{50pt}p{45pt}l||p{50pt}p{48pt}p{70pt}||p{60pt}p{55pt}|}
\hline
& \multicolumn{3}{c||}{Feature-Related} & \multicolumn{3}{c||}{General} & \multicolumn{2}{c|}{Query-Related} \\
& Multiplicity  & Levels  & Representations & Scoring  & Alignment & Dimensionality & Type & Multiplicity  \\ \hline
Bahdanau et al. \cite{bahdanau2014neural} & Singular & Single-Level & Single-Representational  & Additive & Global & Single-Dimensional & Basic & Singular\\ \hline
Luong et al. \cite{luong2015effective} &  Singular & Single-Level & Single-Representational  &  Multiplicative, \newline Location & Global, Local & Single-Dimensional & Basic  & Singular  \\ \hline
Xu et al. \cite{xu2015show} &  Singular & Single-Level & Single-Representational & Additive & Soft, Hard & Single-Dimensional & Basic & Singular  \\ \hline
Lu et al. \cite{NIPS2016_6202} &  Parallel \newline Co-attention & Hierarchical & Single-Representational  &  Additive  &  Global &   Single-Dimensional  & Specialized & Singular  \\ \hline
Yang et al. \cite{yang-etal-2016-hierarchical} & Singular & Hierarchical & Single-Representational  &  Additive  &    Global    &   Single-Dimensional & Self-Attentive &  Singular\\ \hline
Li et al. \cite{li2018hierarchical} &  Singular & Hierarchical & Single-Representational &   Additive    &  Global       &     Single-Dimensional &  Self-Attentive & Singular\\ \hline
Vaswani et al. \cite{vaswani2017attention} &  Singular & Single-Level & Single-Representational &   Scaled-Multiplicative   &  Global       &     Single-Dimensional & Self-Attentive +\newline Basic & Multi-Head +\newline Multi-Hop  \\ \hline
Wallaart and Frasincar \cite{wallaart2019hybrid} &  Rotatory & Single-Level & Single-Representational &   Activated \newline General  &  Global       &     Single-Dimensional  & Specialized & Multi-Hop\\ \hline
Kiela et al. \cite{kiela2018dynamic} & Singular & Single-Level & Multi-Representational &   Additive &  Global       &     Single-Dimensional & Self-Attentive & Singular \\ \hline
Shen et al. \cite{shen2018disan} & Singular & Single-Level & Single-Representational &   Additive &  Global       &     Multi-Dimensional & Self-Attentive & Singular\\ \hline
Zhang et al. \cite{zhang2018selfattention} & Singular & Single-Level & Single-Representational &   Multiplicative &  Global       &     Single-Dimensional & Self-Attentive & Singular\\ \hline
Li et al. \cite{li2019beyond} & Parallel \newline Co-attention & Single-Level & Single-Representational & Scaled-Multiplicative &  Global   &     Single-Dimensional & Self-Attentive +\newline Specialized & Singular\\ \hline
Yu et al. \cite{yu2018qanet} & Parallel \newline Co-attention & Single-Level & Single-Representational & Multiplicative &  Global   & Single-Dimensional  & Self-Attentive +\newline Specialized & Multi-Head\\ \hline
Wang et al. \cite{wang-etal-2019-aspect} & Parallel \newline Co-attention & Single-Level & Single-Representational & Additive &  Reinforced  & Single-Dimensional  & Specialized & Singular\\ \hline
Oktay et al. \cite{oktay2018attention} & Singular & Single-Level & Single-Representational & Additive &  Global   &     Multi-Dimensional & Self-Attentive +\newline Specialized & Singular\\ \hline
Winata et al. \cite{winata2019learning} & Singular & Single-Level & Multi-Representational & Additive &  Global   &  Single-Dimensional & Self-Attentive & Multi-Head\\ \hline
Wang et al. \cite{10.1145/3178876.3186015} & Singular & Single-Level & Single-Representational & Multiplicative &  Global   &  Single-Dimensional & Self-Attentive & Capsule-Based\\ \hline
\end{tabular}
}
\end{center}
\label{Table:PapersByTaxonomy}
\end{table*}

\section{Evaluation of Attention Models}\label{sec:Evaluation}
In this section, we present various types of evaluation for attention models. Firstly, one can evaluate the structure of attention models using the taxonomy presented in Section \ref{sec:Taxonomy}. For such an analysis, we consider the attention mechanism categories (see Fig. \ref{fig:Taxonomy}) as orthogonal dimensions of a model. The structure of a model can be analyzed by determining which mechanism a model uses for each category. Table \ref{Table:PapersByTaxonomy} provides an overview of attention models found in the literature with a corresponding analysis based on the attention mechanisms the models implement.

Secondly, we discuss various techniques for evaluating the performance of attention models. The performance of attention models can be evaluated using \textbf{extrinsic} or \textbf{intrinsic} performance measures, which are discussed in Subsections \ref{sec:Extrinsic} and \ref{sec:Intrinsic}, respectively.

\subsection{Extrinsic Evaluation}\label{sec:Extrinsic}

In general, the performance of an attention model is measured using \textbf{extrinsic performance measures}. For example, performance measures typically used in the field of natural language processing are the BLEU \cite{papineni2002bleu}, METEOR \cite{banerjee2005meteor}, and Perplexity \cite{sennrich2012perplexity} metrics. In the field of audio processing, the Word Error Rate \cite{popovic2007word} and Phoneme Error Rate \cite{schwarz2004towards} are generally employed. For general classification tasks, error rates, precision, and recall are generally used. For computer vision tasks, the PSNR \cite{turaga2004no}, SSIM \cite{ndajah2010ssim}, or IoU \cite{rahman2016optimizing} metrics are used. Using these performance measures, an attention model can either be compared to other state-of-the-art models, or an ablation study can be performed. If possible, the importance of the attention mechanism can be tested by replacing it with another mechanism and observing whether the overall performance of the model decreases \cite{li2019beyond, chen2020residual}. An example of this is replacing the weighted average used to produce the context vector with a simple unweighted average and observing whether there is a decrease in overall model performance \cite{seo2016bidirectional}. This ablation method can be used to evaluate whether the attention weights can actually distinguish important from irrelevant information.

\subsection{Intrinsic Evaluation}\label{sec:Intrinsic}

Attention models can also be evaluated using attention-specific \textbf{intrinsic performance measures}. In \cite{luong2015effective}, the attention weights are formally evaluated via the Alignment Error Rate (AER) to measure the accuracy of the attention weights with respect to annotated attention vectors. \cite{liu2017exploiting} incorporates this idea into an attention model by supervising the attention mechanism using gold attention vectors. A joint loss function consisting of the regular task-specific loss and the attention weights loss function is constructed for this purpose. The gold attention vectors are based on annotated text data sets where keywords are hand-labelled. However, since attention is inspired by human attention, one could evaluate attention models by comparing them to the attention behaviour of humans.

\subsubsection{Evaluation via Human Attention}\label{sec:HumanAttention}

In \cite{liu2017attention}, the concept of \textbf{attention correctness} is proposed, which is a quantitative intrinsic performance metric that evaluates the quality of the attention mechanism based on actual human attention behaviour. Firstly, the calculation of this metric requires data that includes the attention behaviour of a human. For example, a data set containing images with the corresponding regions that a human focuses on when performing a certain task, such as image captioning. The collection of regions focused on by the human is referred to as the ground truth region. Suppose an attention model attends to the $n_f$ feature vectors $\bm f_1, \dots, \bm f_{n_f} \in \mathbb{R}^{d_f}$. Feature vector $\bm f_i$ corresponds to region $R_i$ of the given image, for $i=1,\dots,n_f$. We define the set $G$ as the set of regions that belong to the ground truth region, such that $R_i \in G$ if $R_i$ is part of the ground truth region. The attention model calculates the attention weights $a_1, \dots , a_{n_f}\in \mathbb{R}^{1}$ via the usual attention process. The Attention Correctness ($AC$) metric can then be calculated using (\ref{equation:AttentionCorrectness}).
\begin{equation}\label{equation:AttentionCorrectness}
    \def\sss{\scriptscriptstyle}
    \setstackgap{L}{8pt}
    \def\stacktype{L}
    \stackunder{AC}{\sss 1 \times 1} = \sum_{i: R_i \in G} \stackunder{a_i}{\sss 1 \times 1}.
\end{equation}
Thus, this metric is equal to the sum of the attention weights for the ground truth regions. Since the attention weights sum up to 1 due to, for example, a softmax alignment function, the $AC$ value will be a value between 0 and 1. If the model attends to only the ground truth regions, then $AC$ is equal to 1, and if the attention model does not attend to any of the ground truth regions, $AC$ will be equal to 0.

In \cite{das2017human}, a rank correlation metric is used to compare the generated attention weights to the attention behaviour of humans. The conclusion of this work is that attention maps generated by standard attention models generally do not correspond to human attention. Attention models often focus on much larger regions or multiple small non-adjacent regions. As such, a technique to improve attention models is to allow the model to learn from human attention patterns via a joint loss of the regular loss function and an attention weight loss function based on the human gaze behaviour, similarly to how annotated attention vectors are used in \cite{liu2017exploiting} to supervise the attention mechanism. \cite{liu2017attention} proposes to use human attention data to supervise the attention mechanism in such a manner. Similarly, a state-of-the-art video captioning model is proposed in \cite{yu2017supervising} that learns from human gaze data to improve the attention mechanism.

\subsubsection{Manual Evaluation}\label{sec:Manual}

A method that is often used to evaluate attention models is the manual inspection of attention weights. As previously mentioned, the attention weights are a direct indication of which parts of the data the attention model finds most important. Therefore, observing which parts of the inputs the model focuses on can be helpful in determining if the model is behaving correctly. This allows for some interpretation of the behaviour of models that are typically known to be black boxes. However, rather than checking if the model focuses on the most important parts of the data, some use the attention weights to determine which parts of the data are most important. This would imply that attention models provide a type of explanation, which is a subject of contention among researchers. Particularly, in \cite{jain2019attention}, extensive experiments are conducted for various natural language processing tasks to investigate the relation between attention weights and important information to determine whether attention can actually provide meaningful explanations. In this paper titled ``Attention is not Explanation", it is found that attention weights do not tend to correlate with important features. Additionally, the authors are able to replace the produced attention weights with completely different values while keeping the model output the same. These so-called ``adversarial" attention distributions show that an attention model may focus on completely different information and still come to the same conclusions, which makes interpretation difficult. Yet, in another paper titled ``Attention is not not Explanation" \cite{wiegreffe-pinter-2019-attention}, the claim that attention is not explanation is questioned by challenging the assumptions of the previous work. It is found that the adversarial attention distributions do not perform as reliably well as the learned attention weights, indicating that it was not proved that attention is not viable for explanation.

In general, the conclusion regarding the interpretability of attention models is that researchers must be extremely careful when drawing conclusions based on attention patterns. For example, problems with an attention model can be diagnosed via the attention weights if the model is found to focus on the incorrect parts of the data, if such information is available. Yet, conversely, attention weights may only be used to obtain plausible explanations for why certain parts of the data are focused on, rather than concluding that those parts are significant to the problem \cite{wiegreffe-pinter-2019-attention}. However, one should still be cautious as the viability of such approaches can depend on the model architecture \cite{mohankumar-etal-2020-towards}.

\section{Conclusion}\label{sec:Conclusion}
In this survey, we have provided an overview of recent research on attention models in deep learning. Attention mechanisms have been a prominent development for deep learning models as they have shown to improve model performance significantly, producing state-of-the-art results for various tasks in several fields of research. We have presented a comprehensive taxonomy that can be used to categorize and explain the diverse number of attention mechanisms proposed in the literature. The organization of the taxonomy was motivated based on the structure of a task model that consists of a feature model, an attention model, a query model, and an output model. Furthermore, the attention mechanisms have been discussed using a framework based on queries, keys, and values. Last, we have shown how one can use extrinsic and intrinsic measures to evaluate the performance of attention models, and how one can use the taxonomy to analyze the structure of attention models.

The attention mechanism is typically relatively simple to understand and implement and can lead to significant improvements in performance. As such, it is no surprise that this is a highly active field of research with new attention mechanisms and models being developed constantly. Not only are new mechanisms consistently being developed, but there is also still ample opportunity for the exploration of existing mechanisms for new tasks. For example, multi-dimensional attention \cite{shen2018disan} is a technique that shows promising results and is general enough to be implemented in almost any attention model. However, it has not seen much application in current works. Similarly, multi-head attention \cite{vaswani2017attention} is a technique that can be efficiently parallelized and implemented in practically any attention model. Yet, it is mostly seen only in Transformer-based architectures. Lastly, similarly to how \cite{wallaart2019hybrid} combines rotatory attention with multi-hop attention, combining multi-dimensional attention, multi-head attention, capsule-based attention, or any of the other mechanisms presented in this survey may produce new state-of-the-art results for the various fields of research mentioned in this survey.

This survey has mainly focused on attention mechanisms for supervised models, since these comprise the largest proportion of the attention models in the literature. In comparison to the total amount of research that has been done on attention models, research on attention models for semi-supervised learning \cite{thekumparampil2018attention, nie2018asdnet} or unsupervised learning \cite{mejjati2018unsupervised, he2017unsupervised} has received limited attention and has only become active recently. Attention may play a more significant role for such tasks in the future as obtaining large amounts of labeled data is a difficult task. Yet, as larger and more detailed data sets become available, the research on attention models can advance even further. For example, we mentioned the fact that attention weights can be trained directly based on hand-annotated data \cite{liu2017exploiting} or actual human attention behaviour \cite{liu2017attention, yu2017supervising}. As new data sets are released, future research may focus on developing attention models that can incorporate those types of data.

While attention is intuitively easy to understand, there still is a substantial lack of theoretical support for attention. As such, we expect more theoretical studies to additionally contribute to the understanding of the attention mechanisms in complex deep learning systems. Nevertheless, the practical advantages of attention models are clear. Since attention models provide significant performance improvements in a variety of fields, and as there are ample opportunities for more advancements, we foresee that these models will still receive significant attention in the time to come.


%


\bibliographystyle{IEEEtran}
\bibliography{references}

\vspace{-10pt}

\begin{IEEEbiography}[{\includegraphics[width=1in,height=1.5in,clip,keepaspectratio]{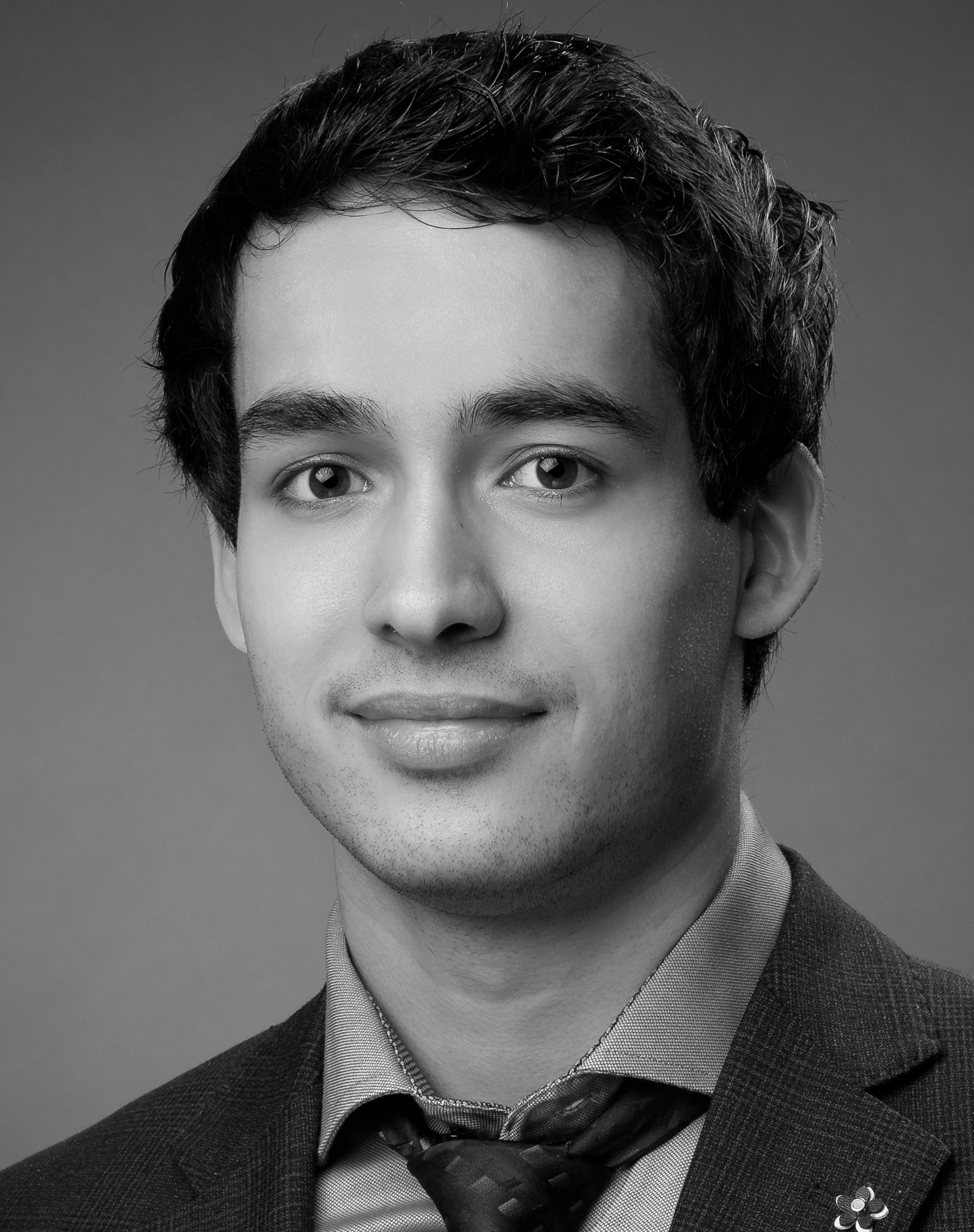}}]{Gianni Brauwers} was born in Spijkenisse, the Netherlands, in 1998. He received the B.S. degree in econometrics and operations research from Erasmus University Rotterdam, Rotterdam, the Netherlands, in 2019, and is currently pursuing the M.S. degree in econometrics and management science at Erasmus University Rotterdam.

He is a Research Assistant at Erasmus University Rotterdam, focusing his research on neural attention models and sentiment analysis.
\end{IEEEbiography}

\vspace{-10pt}

\begin{IEEEbiography}[{\includegraphics[width=1in,height=1.25in,clip,keepaspectratio]{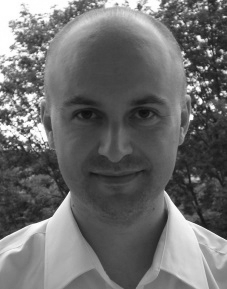}}]{Flavius Frasincar} was born in Bucharest, Romania, in 1971. He 
received the M.S. degree in computer science, in 1996, and the M.Phil. 
degree in computer science, in 1997, from Politehnica University of 
Bucharest, Bucharest, Romania, and the P.D.Eng. degree in computer 
science, in 2000, and the Ph.D. degree in computer science, in 2005, 
from Eindhoven University of Technology, Eindhoven, the Netherlands.

Since 2005, he has been an Assistant Professor in computer science at Erasmus 
University Rotterdam, Rotterdam, the Netherlands. He has published in 
numerous conferences and journals in the areas of databases, Web 
information systems, personalization, machine learning, and the Semantic 
Web. He is a member of the editorial boards of Decision Support Systems, 
International Journal of Web Engineering and Technology, and 
Computational Linguistics in the Netherlands Journal, and 
co-editor-in-chief of the Journal of Web Engineering. Dr. Frasincar is a member of the Association for Computing Machinery.
\end{IEEEbiography}

\vfill




\end{document}